\documentclass{article}

\PassOptionsToPackage{numbers,compress}{natbib}
\usepackage[main, final]{neurips_2026}
\usepackage[normalem]{ulem}
\usepackage[utf8]{inputenc} % allow utf-8 input
\usepackage[T1]{fontenc}    % use 8-bit T1 fonts
\usepackage{xcolor}
\definecolor{linkblue}{HTML}{000080}

\usepackage{hyperref}
\hypersetup{
    colorlinks=true,
    linkcolor=linkblue,
    citecolor=linkblue,
    urlcolor=linkblue
}
\usepackage{url}            % simple URL typesetting
\usepackage{booktabs}       % professional-quality tables
\usepackage{tabularx}       % tables constrained to \linewidth
\usepackage{amsmath}        % display math, \text in math mode, align, etc.
\usepackage{amsfonts}       % blackboard math symbols
\usepackage{nicefrac}       % compact symbols for 1/2, etc.
\usepackage{microtype}      % microtypography
\usepackage[table]{xcolor}  % colors; [table] enables \rowcolor in tabular
\newcommand{\bl}[1]{\textcolor{black!55}{#1}}    % gray text for baseline cells (visual de-emphasis)
\usepackage{graphicx}       % \includegraphics
\graphicspath{{figures/}}   % look for figures in ./figures/
\usepackage{subcaption}     % \subcaption for side-by-side subfigures
\usepackage{wrapfig}        % wraptable / wrapfigure for text-wrap layouts
\usepackage{enumitem}       % itemize/enumerate options like leftmargin=*, nosep
\usepackage{placeins}       % keep appendix floats near their sections
\usepackage[most]{tcolorbox} % styled prompt boxes in the appendix
\tcbuselibrary{listings,breakable}
\newcolumntype{Y}{>{\raggedright\arraybackslash}X}
\lstdefinestyle{appendixlisting}{
  basicstyle=\ttfamily\scriptsize,
  breaklines=true,
  breakatwhitespace=false,
  columns=fullflexible,
  keepspaces=true,
  showstringspaces=false,
  upquote=true,
}
\lstdefinestyle{appendixlistingtiny}{
  basicstyle=\ttfamily\tiny,
  breaklines=true,
  breakatwhitespace=false,
  columns=fullflexible,
  keepspaces=true,
  showstringspaces=false,
  upquote=true,
}
\tcbset{
  appendixbox/.style={
    enhanced,
    colback=gray!5!white,
    colframe=gray!75!black,
    fonttitle=\bfseries,
    rounded corners,
    width=\linewidth,
    breakable,
    boxsep=1mm,
    left=1mm,
    right=1mm,
    top=1mm,
    bottom=1mm,
    before skip=0.75em,
    after skip=0.75em,
  },
  promptbox/.style={
    appendixbox,
    listing only,
    listing options={style=appendixlisting},
  },
  samplebox/.style={
    enhanced,
    colback=gray!5!white,
    colframe=gray!75!black,
    fonttitle=\bfseries,
    rounded corners,
    width=\linewidth,
    boxsep=0.75mm,
    left=0.75mm,
    right=0.75mm,
    top=0.75mm,
    bottom=0.75mm,
    before skip=0.75em,
    after skip=0.75em,
    listing only,
    listing options={style=appendixlistingtiny},
  },
  longpromptbox/.style={
    appendixbox,
    listing only,
    listing options={style=appendixlistingtiny},
  },
}

\title{\raisebox{-0.3\height}{\includegraphics[height=1.6em]{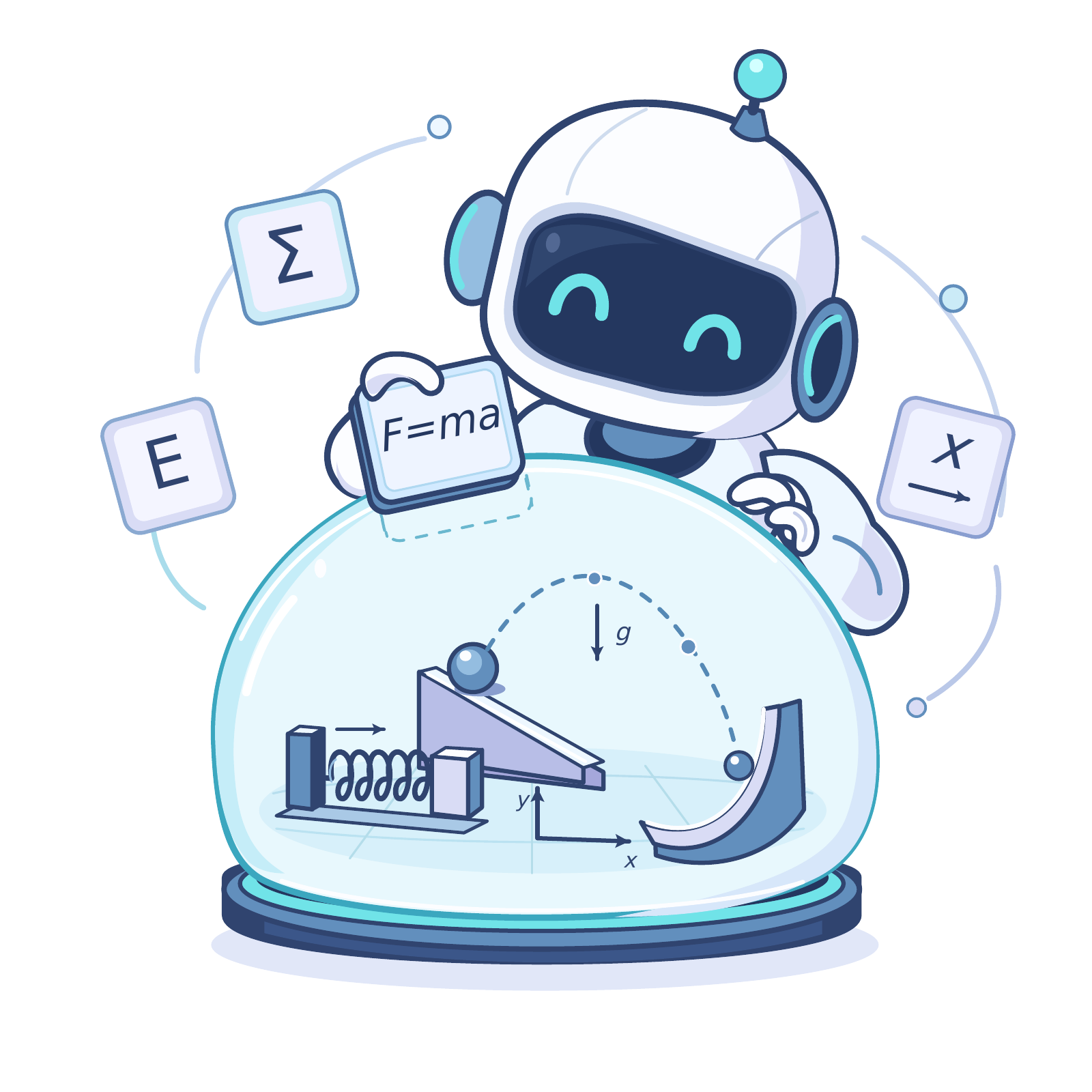}}\hspace{0.35em}Axiomatic World Modeling for Physics Reasoning}

\author{%
  Xinye Yang$^{1}$ \quad
  Zhenyang Liu$^{1}$ \quad
  Yuxuan Wang$^{2,\dagger}$ \quad
  Yuanyuan Lei$^{3,\dagger}$ \\
  $^{1}$Northwestern University, Evanston, IL \\
  $^{2}$Department of Physics, University of Florida, Gainesville, FL \\
  $^{3}$Computer \& Information Science and Engineering,
  University of Florida, Gainesville, FL \\[0.35em]
  \small
  \textbf{Project Website:}
  \url{https://lei-nlp-lab.github.io/physics_reasoning_neurips_2026/}
}
\begin{document}

\maketitle
{\renewcommand{\thefootnote}{\fnsymbol{footnote}}
 \footnotetext[2]{Corresponding authors: \texttt{yuanyuan.lei@ufl.edu}, \texttt{yuxuan.wang@ufl.edu}}}

\begin{abstract}
  Recent research-level physics-reasoning benchmarks for LLMs~\citep{pan2026cmtbenchmarkbenchmarkcondensedmatter} reveal a failure mode that lies not in knowledge or execution, but in modeling the world being reasoned about.  Models often silently alter given quantities, fabricate laws, or inject unstated assumptions to force a problem into familiar patterns, and then reason inside this corrupted world. We define this premise-level divergence as \textbf{modeling drift}: the gap between a model's internal axiomatic world model (AWM) and the true physical world specified in the problem. Although RLVR and PRM have yielded substantial gains in mathematics and code, their reward shapes do not directly supervise world modeling in axiomatic domains such as physics. We propose \textbf{\textit{Reinforcement Learning with Axiomatic World Modeling}} (RLAWM), which optimizes the policy by penalizing divergence between its inferred axiomatic world model and the true physics world. RLAWM probes physics reasoning at two hierarchical levels (modeling and solving), each scored by a physically grounded reward. The \textbf{modeling reward} targets abstract knowledge formulation, evaluating whether the model captures the true axiomatic world before attempting calculation. The \textbf{solving reward} targets instantiated knowledge reasoning, checking whether the model's concrete derivation remains consistent with the axiomatic world. This two-level structure separates \textit{what world the model believes it is reasoning in} from \textit{how it reasons within that world}. The framework operates exclusively during training, with no extra cost during inference. RLAWM improves performance by up to $+12.2$ points on PhysReason and $+5.2$ on PHYSICS over the baseline.  It also generalizes zero-shot across difficulty, source, and domain, with a $2.45{\times}$ gain and $+10$ points over the strongest baseline on olympiad-level physics and math \footnote{The link for code and model is:  \url{https://github.com/lei-nlp-lab/physics_reasoning_neurips_2026}.}.
  
\end{abstract}

%% ============================================================
%% §1  Introduction
%% ============================================================
\section{Introduction}
\label{sec:intro}

\begin{figure}[!htbp]
  \centering
  \includegraphics[width=1.0\linewidth]{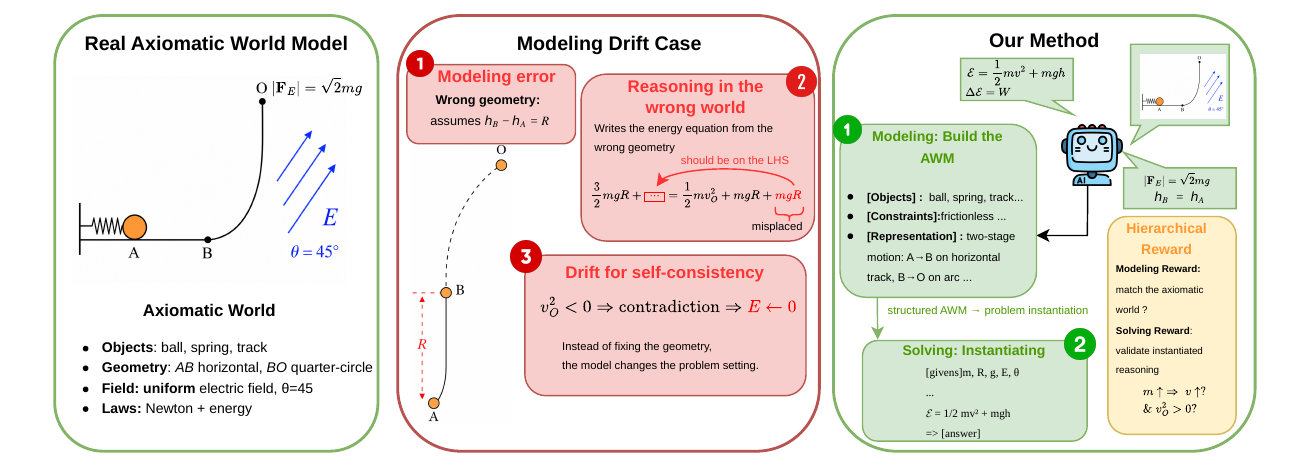}
  \caption{\textbf{Modeling drift on a representative physics problem.}
    \textbf{Left:} the true axiomatic world model (AWM): objects, geometry, field, and laws.
    \textbf{Middle:} a representative drift trajectory of the base model: wrong geometry
    propagates into a malformed energy equation, and upon deriving $v_O^2<0$ the model
    silently sets $E\!\leftarrow\!0$ for self-consistency instead of repairing the geometry.
    \textbf{Right:} an RLAWM-trained model recovers the correct geometry and reasons
    consistently within the true AWM.}
  \label{fig:overview}
\end{figure}

Despite great progress in LLMs' reasoning capabilities, recent benchmarks \citep{pan2026cmtbenchmarkbenchmarkcondensedmatter, wang2025cmphysbenchbenchmarkevaluatinglarge, zhu2025probingcriticalpointcritpt} have revealed that state-of-the-art LLMs still struggle on research-level physics problems. CMT-Benchmark~\citep{pan2026cmtbenchmarkbenchmarkcondensedmatter} identifies several recurring failure modes. Some point to the absence of a coherent internal model of the physical setup: models manipulate equations fluently without grounding them in the underlying lattice or momentum-space geometry, default to common textbook patterns instead of solving the specific problem given, and miss mappings to known solvable cases. Others reflect a failure to stay faithful to the problem as stated: models silently add unstated assumptions (typically a conservation law that wasn't given) to reduce hard problems to familiar ones, and produce final expressions that violate basic constraints such as symmetry, scaling, or dimensional consistency. These failures lie in the model of the problem and the physical world implicitly constructed by it, not in the algebra.

Compared with math problems, which typically benefit from a formal structure of axioms, theorems, and goals, a physics problem arrives underspecified. The solver has to first model the situation: decide which physical laws apply, which effects can be neglected, what limiting case admits a tractable solution, and what invariants the answer must respect. Errors at this modeling stage are not deductive errors; they are errors about which problem is being solved. We dub this modeling process \textbf{axiomatic world modeling (AWM)}. These failure modes are prevalent even in much simpler physics problems. 
Figure~\ref{fig:overview} contrasts the true axiomatic world model of a
representative problem from PhysReason~\citep{zhang2025physreasoncomprehensivebenchmarkphysicsbased} (left) with the drifted version that current language models
silently believe while solving it (middle).
A spring-released ball slides along a horizontal rail AB to point B, then descends
along a quarter-arc BO into a region with a uniform electric field.
This problem rests on a small set of axioms: the geometry (AB horizontal, so $h_A = h_B$),
the magnitude and direction of the field, the conservation laws, and the initial state.
Under this AWM, correct solution simply respects the conditions and constraints and does the energy bookkeeping.
Yet strong language models routinely corrupt this world model during reasoning and continue
inside the corrupted version.
The corruption typically unfolds in three steps.
\textbf{(1) Modeling error.}
The model writes the geometry as $h_B - h_A = R$, contradicting the problem's
``AB horizontal'' clause which propagates to downstream calculation.
\textbf{(2) Reasoning inside the wrong world.} The model then applies an incorrect version of work-energy relation,  obtaining the unphysical result
$v_O^2 = -gR$.
\textbf{(3) Drift for self-consistency.}
Faced with this physical impossibility, the model does not return to question its premise;
it rewrites the problem instead, declaring $E = 0$ so the contradiction disappears. The failure is not simply in carrying out the calculation, but in what the model silently believes about the physical world.
%We call this premise-level failure mode \textbf{modeling drift}: the divergence between a model's inferred AWM and the true physics world specified in the problem. Modeling drift appears systematic across physics benchmarks~\mbox{\citep{zhang2025physreasoncomprehensivebenchmarkphysicsbased}}. Unlike general factual hallucination~\citep{ji2023hallucination}, it occurs at the premise level: \textbf{the model reasons coherently inside a wrong physics world.}
We term this premise-level failure \textbf{modeling drift}: the divergence between a model's inferred AWM and the true physics world specified in the problem. This failure mode has been observed across physics reasoning benchmarks~\mbox{\citep{zhang2025physreasoncomprehensivebenchmarkphysicsbased, feng2025physicsbenchmarkingfoundationmodels}}. Unlike general factual hallucination~\citep{ji2023hallucination}, modeling drift occurs at the premise level: \textbf{the model reasons coherently inside a wrong physics world.}

Existing training and verification methods leave this modeling drift failure largely unaddressed. RLVR~\citep{deepseek2025r1, lambert2024tulu3} and process reward models (PRMs)~\citep{lightman2023verify, wang2024mathshepherd} have produced strong gains in math and code reasoning, domains where the premises are typically given in near-formal form and the modeling layer is comparatively shallow. Physics presents a different challenge: the solver must first infer the relevant axiom world before deduction can begin. Standard reward signals rarely supervise this step. Outcome-only rewards evaluate only the final answer, while step-level signals such as PRM~\citep{luo2024omegaprm} and LLM-as-Judge~\citep{zheng2023judge} primarily assess the textual plausibility or local correctness of individual reasoning steps. As a result, they can miss whether the policy has instantiated the correct physical world model in the first place. We analyze this gap and the closest related reward designs in \S\ref{sec:related_work}.

The key opportunity for reducing modeling drift comes from an asymmetry unique to axiomatic domains: \textbf{verifying a physical claim is often easier than deriving it}. In physics, checking whether a derivation violates a conservation law or dimensional homogeneity can be straightforward, whereas deriving the result from scratch is substantially harder. Carrying this idea over to language models is challenging because verification is inherently interactive, where we can run the instance, perturb its inputs, and check the invariants, whereas an LLM’s reasoning trace is usually only static text that can be read but not interactively probed. Our core insight is therefore to convert the reasoning trace into an executable axiomatic world model, where we can verify it automatically and each interaction test yields a training signal that can be fed back to shape the world model underlying the reasoning.

Building on this asymmetry, we propose \textbf{\textit{Reinforcement Learning with Axiomatic World Modeling}} (RLAWM), which probes physical reasoning at two hierarchical levels (modeling and solving) that separate \textit{what world the policy believes it is reasoning in} from \textit{how it reasons within that world}. The \textbf{Modeling reward} targets abstract knowledge formulation, by constructing the abstract representation of the axiomatic physics world and scoring this inferred world representation against the true physics world specified in the problem. The \textbf{Solving reward} targets instantiated knowledge reasoning, by materializing the axiomatic world model into an executable instance and probing its behavior via interactive checkers, such as tests of physical consistency under controlled perturbations. Together, the two rewards provide complementary supervision for AWM: the modeling reward directly supervises the inferred axiomatic world model and teaches the policy which world to construct from the problem statement; while the solving reward provides instantiated supervision on the reasoning process by testing how that world behaves when executed, and this instantiated feedback flows back to shape the abstract world modeling that governs the reasoning. In this sense, the modeling reward anchors world construction, while the solving reward refines that world through instantiated interaction.

RLAWM's gains hold consistently across three evaluation regimes. First, on PhysReason~\citep{zhang2025physreasoncomprehensivebenchmarkphysicsbased} and PHYSICS~\citep{feng2025physicsbenchmarkingfoundationmodels}, across two model families and three scales, RLAWM achieves the strongest Overall score for every backbone and shows substantial gains across difficulty levels and most subdomains. Second, a sharper separation appears on the SubPass@$k$ curve. SubPass@$k$ takes the best of $k$ samples per problem and tracks the reasoning boundary the policy can reach. \citet{yue2025rlvr_limit} report that RLVR and PRM raise SubPass@$1$ but regress toward the base model at large $k$, a phenomenon they call \textbf{diversity collapse}: the policy tightens around the base model's existing capabilities rather than enlarging them. In contrast, RLAWM optimizes reasoning against world-model constraints and physical invariants, rather than relying only on answer matching or step-level guidance. This encourages the policy to explore solutions that remain physically valid, allowing performance to continue improving as $k$ grows. Third, under out-of-distribution evaluation, RLAWM also generalizes better than the baselines along three axes of distribution shift: across difficulty levels on OlympiadBench, across data sources on PHYBench, and across domains on AIME~2026. Taken together, these results show that RLAWM not only improves accuracy, but also expands the reasoning boundary, and transfers more effectively beyond the training distribution.

%% ============================================================
%% §2  Related Work
%% ============================================================
\section{Related Work}
\label{sec:related_work}

\textbf{RL Reward Signals Do Not Address Modeling Drift.} Existing RL methods for LLM reasoning have yielded strong gains in mathematics and code, but their rewards rarely supervise the modeling level required for physics. Outcome-only RLVR~\citep{deepseek2025r1, lambert2024tulu3, shao2024deepseekmath} and free-form answer verification~\citep{ma2026generalreasoner, su2025crossingrewardbridgeexpanding} only check the conclusion. They can therefore give false-positive rewards when a correct answer comes from wrong premises. Step-level methods such as PRMs~\citep{lightman2023verify, wang2024mathshepherd}, LLM-as-Judge~\citep{zheng2023judge}, and Rubrics-as-Rewards~\citep{gunjal2025rar} score intermediate steps and give denser supervision. However, they mainly judge whether each step reads plausibly or is locally correct. Text alone is also an unreliable signal, since a chain-of-thought rationale may not reflect how the model actually reached its answer~\citep{turpin2023unfaithful, lanham2023faithfulness}. \textsc{LogicReward}~\citep{xu2026logicrewardincentivizingllmreasoning} goes beyond text and verifies each step with a theorem prover. Yet it judges soundness only against the axioms the policy has already adopted. A sound derivation inside a drifted world can therefore still be rewarded. \citet{yue2025rlvr_limit} further show that outcome- and step-level rewards suffer from diversity collapse in mathematics and code. \citet{wang2025dph_divergence} trace part of this collapse to the divergence term in the RL objective. Overall, these reward signals assess final-answer correctness, surface-level textual features, or local consistency, but none directly evaluates whether the policy has constructed an accurate model of the underlying physical scenario. RLAWM addresses this gap by decomposing physics problem solving into a hierarchical process of modeling and reasoning. Specifically, it introduces a modeling reward that incentivizes faithful representation of the physical setup and a solving reward that guides the policy's reasoning within the constructed physical model.

%% ============================================================
%% §3  Reinforcement Learning with Axiomatic World Modeling
%% ============================================================
\section{Reinforcement Learning with Axiomatic World Modeling}
\label{sec:method}

\begin{figure}[!t]
  \centering
  \includegraphics[width=1.0\linewidth]{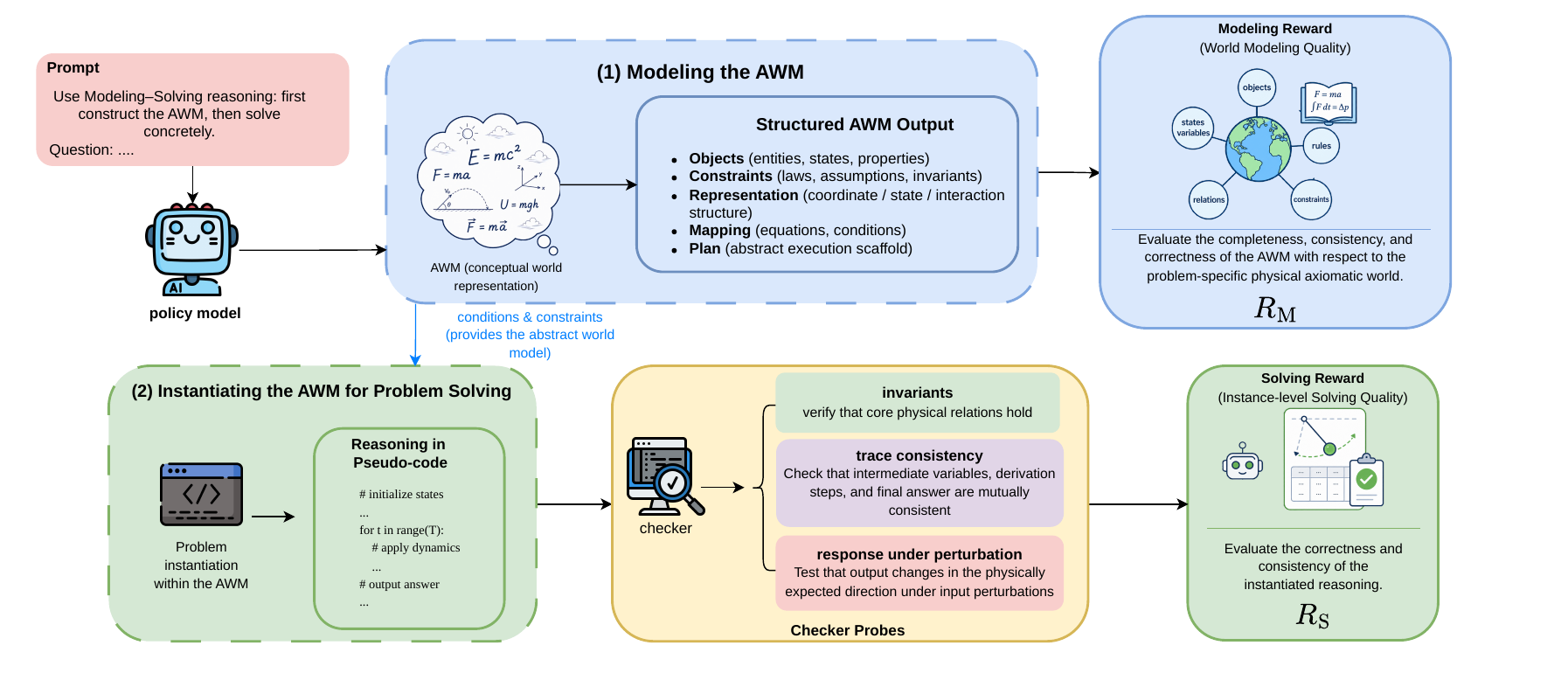}
  \caption{\textbf{The RLAWM pipeline.}
    RLAWM probes physical reasoning at two hierarchical levels, each carrying a grounded reward.
    \textbf{Modeling (top)} concerns how the policy builds its world.
    It externalizes the AWM as a five-field \texttt{[modeling]} block,
    which $R_M$ scores against the problem's ground-truth AWM.
    \textbf{Solving (bottom)} concerns how the policy executes reasoning over that AWM:
    it writes its reasoning as pseudocode, where a frozen checker probes for physical
    invariants, trace consistency, and response under perturbations to yield
    $R_S$.}
  \label{fig:pipeline}
\end{figure}

\subsection{Physical Reasoning through a Belief-State Lens}
\label{sec:method_overview}

Figure~\ref{fig:pipeline} summarizes the RLAWM pipeline, which reduces modeling drift by aligning the policy's internal axiomatic world model (AWM) with the problem's true axiom set. Because this AWM is implicit in the policy’s reasoning chain and parameter weights, it cannot be read or optimized directly. RLAWM therefore forces the policy to externalize its belief at two hierarchical levels, separating \textit{what world the policy believes it is reasoning in} from \textit{how it reasons within that world}.

The two rewards in RLAWM are conceptually complementary. The \textbf{modeling reward} ($R_M$) targets abstract knowledge formulation: it constructs an abstract representation of the policy’s inferred AWM and scores it against the true axiom set specified by the problem. This reward directly anchors the policy to the correct physical premises. The model is also expected to construct an effective axiomatic world \texttt{representation} that supports faithful and tractable reasoning. As the example in Appendix~\ref{app:case_studies} shows, multiple structural commitments, such as a timeline or an interaction graph, may all be physically valid, but they are not equally useful for downstream deduction. A poorly chosen world representation can make the derivation unnecessarily difficult and thereby increase the likelihood of execution errors. Both the premises and the representation the policy commits to therefore shape all downstream reasoning.

The \textbf{solving reward} ($R_S$) then targets instantiated knowledge reasoning by materializing the AWM into an executable instance and probing its behavior via interactive checkers, such as tests of physical consistency under controlled perturbations. This provides instantiated supervision on the reasoning execution process. Crucially, this instantiated feedback also reversely shapes abstract world modeling: by penalizing derivations that become intractable or inconsistent, $R_S$ teaches the policy to construct the most effective axiomatic world \texttt{representation} for a given physics problem.

To make both rewards computable, we guide the policy to split its output into two corresponding blocks: a \texttt{[modeling]} block that externalizes the AWM, and a \texttt{[solving]} block that expresses the reasoning trajectory as pseudocode. $R_M$ scores the \texttt{[modeling]} block to catch premise errors before they propagate, while $R_S$ probes the \texttt{[solving]} block to catch execution inconsistencies. Appendix~\ref{app:notation} summarizes the notation; Appendix~\ref{app:belief} expands on the belief-state framing. Both rewards are designed solely for training and require no external inference-time support.

\subsection{Modeling Reward: What World Does the Policy Believe In?}
\label{sec:modeling_reward}

\paragraph{Externalize the policy's implicit AWM in the \texttt{[modeling]} block.}
The AWM is the policy's believed axiomatic setup during reasoning, implicit
in its reasoning chain, so a reward cannot score it directly.
To compare the AWM with the ground-truth axiom set, the policy externalizes it
as $\hat{m}$, the full \texttt{[modeling]} block. This block has five fields.
\texttt{objects} and \texttt{constraints} record the problem's explicit and
implicit entities, their properties, and the physical laws involved.
\texttt{representation} chooses a structural backbone, as a human physicist
would: \texttt{coordinate}, \texttt{timeline}, \texttt{interaction\_graph},
\texttt{state\_table}, or \texttt{none}.
\texttt{mapping} turns the constraints into governing equations.
\texttt{plan} outlines how to solve each sub-question.
These last two fields also bridge modeling and solving (\S\ref{sec:solving_reward}).
Full schema and per-field examples are in Appendix~\ref{app:modeling_schema}.

\paragraph{Physics Modeling Fidelity (PMF) quantifies the gap between the
externalized AWM and the ground-truth axiom set.}
To reduce reliance on holistic judgments of physical correctness, we
design PMF as a two-step procedure of schema-bounded motif extraction
followed by rule-based aggregation: a frozen
LLM\footnote{Both the motif extractor here and the soft matcher introduced
below use \texttt{gpt-4o-mini-2024-07-18}.} extracts structured motifs from
$\hat{m}$ (the policy's output) and $m^*$ (ground-truth) under the same
schema prompt, and a rule-based function then computes F1 ~\citep{powers2020evaluationprecisionrecallfmeasure} over the matched
motif sets (prompts in Appendix~\ref{app:pmf_prompts}).
This design shifts the LLM from a reasoning judge to a controlled extractor,
so its reasoning biases cannot enter the reward path.
Three motif types target complementary failure modes:
\textbf{QEL} (\texttt{quantity}--\texttt{law}--\texttt{context}) checks
whether the right law covers the right variables;
\textbf{VCL} (Variable Coupling Link) checks whether two equations couple
correctly through a shared variable;
\textbf{AEB} (\texttt{assumption}--\texttt{law}--\texttt{context}) checks
whether each equation's assumptions are present.
We provide detailed explanations and comparisons of the three motif types in Appendix~\ref{app:motif_taxonomy}.
Motifs are matched in two passes: first by rule for exact and normalized
matches, then by an LLM soft matcher for semantically equivalent
but textually different cases (variable renames, unit rewrites, equation
flips). The three F1 scores are then averaged equally:
\[R_M^{\text{PMF}} = \tfrac{1}{3}\bigl(\mathrm{F1}_{\text{QEL}} + \mathrm{F1}_{\text{VCL}} + \mathrm{F1}_{\text{AEB}}\bigr).\]

\subsection{Solving Reward: Did Reasoning Stay Within That World?}
\label{sec:solving_reward}

\paragraph{Verifying Physical Consistency.}
The solving reward verifies the physical consistency of the reasoning carried out within the policy's externalized AWM. Correct abstract modeling alone is insufficient: as illustrated in Appendix~\ref{app:case_studies}, two rollouts with identical abstract AWMs can still yield sharply diverging execution trajectories based on their \texttt{representation} choices. A suboptimal structural commitment often causes the model to stumble during calculation, leading to concrete derivations that violate basic physical laws. To address this, the solving reward acts as an instantiated filter, verifying that the actual execution remains strictly consistent with physics regardless of the chosen path. The \texttt{mapping} and \texttt{plan} fields define the constraints the execution procedure must respect. A  frozen checker then runs a battery of programmatic unit tests on the \textbf{executable instance} of the policy's reasoning trace. Each checker binds problem entities to a shared library of 147 reusable physics-law templates. Appendix~\ref{app:checker} describes the library construction and automated binding pipeline. Appendix~\ref{app:checker_diag} evaluates checker coverage and failure modes. This turns the verification-derivation asymmetry of axiomatic domains~\citep{cobbe2021trainingverifierssolvemath, lightman2023verify, uesato2022process}, into a trainable signal $R_S$ that covers three observable consequences of execution-level inconsistency:
\begin{itemize}[leftmargin=*,nosep]
    \item \textbf{Invariants:} verifies relations such as conservation laws, kinematic equations, and force balance.
    \item \textbf{Trace consistency:} checks that intermediate variables, derivation steps, and the final answer agree.
    \item \textbf{Response under perturbations:} perturbs inputs and checks that outputs move as expected.
\end{itemize}
Each test either passes or fails, and all tests count equally:
\[ R_S = \tfrac{1}{|T|}\textstyle\sum_{t \in T} p_t, \]
where $T$ is the set of tests in the checker and $p_t \in \{0,1\}$ marks a pass. Interface tests, when present, count as ordinary tests.
$R_S$ thus rewards a rollout for reasoning inside the problem's true physical world, regardless of whether its final answer is right. This is the world that $R_M$ trains the policy to construct.
Because each checker uses reusable physical-law templates instantiated with
problem-specific variables and relations,
optimizing $R_S$ does not fit the policy to one particular checker; it internalizes
law-abiding execution into the policy's own reasoning behavior.

\paragraph{Implementation via Pseudocode Translation.}
To generate the \textbf{executable instance} required for verification, we implement a pipeline that translates the policy's reasoning trace into functional code. As introduced in \S\ref{sec:method_overview}, the policy is guided to write its derivation steps in structured pseudocode. This intermediate format serves as a necessary compromise between the strict syntax requirements of direct Python generation (which degrades reasoning quality) and the unparseable ambiguity of natural language. We employ a translator $f$, implemented as a frozen \texttt{Qwen2.5-Coder-1.5B-Instruct}, to map this pseudocode $c$ into the interactively executable Python instance $\hat{w} = f(c)$.
The translator performs two critical roles. First, it preserves the original trace structure (the order of variable definitions, derived-from relations, and control flow). Second, it binds the policy's freely named variables to the fixed fields the checker $g$ expects. If the translator fails to produce code, we drop $R_S$ for this rollout and renormalize the remaining weights. Otherwise, the checker runs $\hat{w}$ in a sandbox, and a $\hat{w}$ that fails to execute receives $R_S = 0$. If it is executable but misses or misnames a required field, that specific unit test fails (contributing $0$ to $R_S$), while the remaining tests proceed. Checker generation and implementation details are in Appendix~\ref{app:checker}. As shown in Appendix~\ref{app:checker_diag}, the checker densifies supervision where the outcome reward is otherwise silent, and its rare failures are diagnosable. Two failure cases there mark where the method's coverage ends.

\subsection{Joint Training Objective}
\label{sec:joint_objective}

We optimize a joint reward that unifies the previously defined modeling ($R_M$) and solving ($R_S$) signals with final-answer correctness ($R_A$): \[
  R_{\text{joint}} = \lambda_M R_M + \lambda_S R_S + \lambda_A R_A
\]

%% ============================================================
%% §6  Experimental Setup
%% ============================================================
\section{Experimental Setup}
\label{sec:setup}

\subsection{Datasets}
\label{sec:datasets}

We evaluate on  five benchmarks that probe complementary axes. For in-distribution evaluation, we use matched $9{:}1$ train/test splits  on two benchmarks. \textbf{PhysReason}~\citep{zhang2025physreasoncomprehensivebenchmarkphysicsbased} contains $1{,}200$ problems spanning knowledge-recall to \emph{difficult} multi-step reasoning. \textbf{PHYSICS-text\_only}~\citep{feng2025physicsbenchmarkingfoundationmodels} adds $996$ university-to-graduate level problems across six subdomains, complementing PhysReason's coverage. For out-of-distribution (OOD) evaluation, policies trained on PhysReason are tested zero-shot  along three axes of distribution shift. Cross-difficulty keeps physics problems in the style of PhysReason but raises their difficulty to the olympiad level, using \textbf{OlympiadBench-Phys}, the physics subset of OlympiadBench~\mbox{\citep{he2024olympiadbenchchallengingbenchmarkpromoting}} ($236$ competition-level problems with verifiable closed-form answers). Cross-source keeps the domain but changes the provenance and modality of the problems. PhysReason collects public exam and competition problems, and most include diagrams. \textbf{PHYBench}~\citep{PHYBENCH_NeurIPS2025} instead contains $500$ original text-only problems written by physics students and unavailable online. Its answers are symbolic expressions, and its difficulty spans high school to olympiad level. Cross-domain leaves physics, using \textbf{AIME~2026}~\citep{dekoninck2026matharena}. For AIME, all methods use the same \emph{Zero-RL} protocol. Here, Zero-RL means applying GRPO directly to the base policy without an SFT cold start. Each method trains only on PhysReason and is then evaluated zero-shot on AIME. No AIME training data or test-time checker is used. We omit physics SFT because it could introduce a domain mismatch between physics training and mathematics evaluation. Full split details are in Table~\ref{tab:datasets} and Appendix~\ref{app:datasets}. The in-distribution splits thus measure fit to the training distribution, and the three OOD axes measure transfer beyond it.
\subsection{Models and Baselines}
\label{sec:baselines}

We use three base policies spanning two model families and three parameter
scales: \textbf{Qwen3-VL-4B-Instruct} and
\textbf{Qwen3-VL-8B-Instruct}~\citep{bai2025qwen3vltechnicalreport}, plus
\textbf{Gemma-3-12B-IT}~\citep{gemmateam2025gemma3technicalreport}. We use multimodal variants of both families because $81\%$ of PhysReason problems
include figures.  These backbones let
us probe within-family scaling (Qwen 4B$\to$8B) and cross-family transfer.

We compare against five baselines. \textbf{Vanilla LLM} uses the base policy without training. \textbf{RLVR}~\citep{deepseek2025r1, lambert2024tulu3} uses a reward that is based only on final-answer correctness. \textbf{Direct-Likert}~\citep{zheng2023judge, kim2024prometheus2opensource} has a frozen judge LLM score each (prompt, response) pair on a $1$--$5$ scale, normalized to $[0,1]$. \textbf{Reference-Likert}~\citep{zheng2023judge, kim2024prometheus2opensource} is the same as Direct-Likert, except that the judge additionally sees the ground-truth reasoning $x^*$. Finally, \textbf{Rubrics as Rewards (RaR)}~\citep{gunjal2025rar} generates a per-prompt rubric and uses a frozen judge to aggregate it into a single reward. These baselines span no training, an outcome-only reward, and three judge-based rewards.

\subsection{Training and Evaluation Protocol}
\label{sec:eval_protocol}

\paragraph{Training protocol.}
Except for the AIME cross-domain experiment described in \S\ref{sec:datasets},
all methods follow the same two-stage paradigm: an SFT cold start followed by
GRPO~\citep{shao2024deepseekmath}, with shared hyperparameters; only the
reward source differs. Baselines do SFT on the dataset's native ground-truth
solutions; RLAWM does SFT on the same solutions restructured into the
\texttt{[modeling]} / \texttt{[solving]} schema described in
\S\ref{sec:method} by \texttt{gpt-4o-mini}\footnote{We use the
\texttt{gpt-4o-mini-2024-07-18} model snapshot throughout.}~\citep{openai2024gpt4technicalreport}.
For AIME, all methods instead apply GRPO directly to the base policy on
PhysReason, without an SFT cold start.
RLAWM additionally needs a checker $g$, bound per problem from the shared
theorem library for every individual training instance, generated before
training and frozen throughout. Prompts, the theorem library, and full
hyperparameters are in Appendices~\ref{app:modeling_schema}, \ref{app:checker},
and~\ref{app:hyperparams}. Any gap between RLAWM and a baseline therefore
traces to the reward or to the SFT restructuring it requires, which
\S\ref{sec:ablation_decouple} separates.
Appendices~\ref{app:hyperparams} and~\ref{app:checker} report the
training compute, wall-clock time, and checker-synthesis cost. All checker
components are training-only and require no external inference-time support.
For the joint reward, we set $\lambda_M = \lambda_S = 0.4$ and $\lambda_A = 0.2$. $R_M$ and $R_S$ share equal weight because they supervise the same AWM at two levels. As shown in Appendix~\ref{app:checker_diag}, $R_A$ catches errors that the checker cannot see, most often a wrong final simplification. We keep its weight small because, as reported in \S\ref{sec:signal_analysis}, the answer reward is sparse and often gives no gradient within a GRPO group.

\paragraph{Evaluation.}
Each problem $q_i$ has $J_i$ sub-questions, and the policy outputs
sub-answers $\hat{a}_{i,j}(y)$ inside \texttt{[ans]}\ldots\texttt{[ans]}
markers. \texttt{deepseek-v4-flash-0423}~\citep{deepseekai2026deepseekv4highlyefficientmilliontoken} judges each prediction against ground truth,
normalizing units and algebra, without seeing the reasoning trace. The
per-response score is the fraction judged correct,
$s(y, q_i) = J_i^{-1}\sum_{j=1}^{J_i} J(\hat{a}_{i,j}(y),\, a^*_{i,j})$.
We adopt \textbf{SubPass@$k$}, a sub-question-level variant of Pass@$k$.
Because $s \in [0,1]$ is fractional, SubPass@$k$ rewards partial credit; it
reduces to Pass@$k$ when $J_i = 1$. \textbf{SubPass@1}
reports $s$ under greedy decoding, averaged over the $N$ test problems;
\textbf{SubPass@$k$} draws $k$ samples per problem at $T = 0.8$ and averages
the best-of-$k$ score:
{\setlength{\abovedisplayskip}{2pt}\setlength{\belowdisplayskip}{2pt}%
\setlength{\abovedisplayshortskip}{0pt}\setlength{\belowdisplayshortskip}{2pt}%
\[
  \mathrm{SubPass}@k = \frac{1}{N} \sum_{i=1}^{N} \max_{j \in [k]} s\!\left(y^{(j)}_i, q_i\right),
  \qquad k \in \{1, 5, 10, 20, 50, 100\}.
\]}

%% ============================================================
%% §7  Main Results
%% ============================================================
\section{Main Results}
\label{sec:results}

\subsection{SubPass@1 Across Baselines, Difficulties, and Subdomains}
\label{sec:pass1}

\begin{table}[!t]
\centering
\caption{\textbf{SubPass@1 (\%) on PhysReason and PHYSICS-text\_only}.
Best per column within each backbone is in \textbf{bold}.
RLAWM's $R_M$ with \texttt{gpt-4o-mini-2024-07-18} as the scorer.}
\label{tab:main_results}
\scriptsize
\setlength{\tabcolsep}{3pt}
\renewcommand{\arraystretch}{1.08}
\resizebox{\linewidth}{!}{%
\begin{tabular}{@{}l ccccc ccccccc@{}}
\toprule
& \multicolumn{5}{c}{\textbf{PhysReason}} & \multicolumn{7}{c}{\textbf{PHYSICS-text\_only}} \\
\cmidrule(lr){2-6}\cmidrule(lr){7-13}
Method & Easy & Medium & Difficult & Knowledge & Overall & Stat.~Mech. & Quantum & Atomic & E\&M & Mechanics & Optics & Overall \\
\midrule
\rowcolor{gray!18}\multicolumn{13}{@{}l}{\textit{Backbone: Qwen3-VL-8B-Instruct}} \\
\bl{Vanilla LLM}        & \bl{39.5} & \bl{15.5} & \bl{13.3} & \bl{46.0} & \bl{30.0} & \bl{16.3} & \bl{17.3} & \bl{ 9.7} & \bl{33.8} & \bl{18.0} & \bl{16.7} & \bl{18.5} \\
\bl{RLVR}               & \bl{39.1} & \bl{22.5} & \bl{17.5} & \bl{47.3} & \bl{32.8} & \bl{16.8} & \bl{16.3} & \bl{14.0} & \bl{25.0} & \bl{16.0} & \bl{36.1} & \bl{19.2} \\
\bl{Direct-Likert}      & \bl{43.8} & \bl{25.7} & \bl{20.6} & \bl{52.8} & \bl{37.1} & \bl{23.7} & \bl{\textbf{23.8}} & \bl{14.4} & \bl{33.8} & \bl{19.7} & \bl{31.6} & \bl{23.8} \\
\bl{Reference-Likert}   & \bl{31.0} & \bl{20.1} & \bl{20.3} & \bl{55.6} & \bl{33.2} & \bl{18.8} & \bl{22.3} & \bl{14.4} & \bl{31.6} & \bl{18.2} & \bl{34.2} & \bl{22.2} \\
\bl{Rubrics as Rewards}                & \bl{38.7} & \bl{24.4} & \bl{19.5} & \bl{50.0} & \bl{34.4} & \bl{19.0} & \bl{23.0} & \bl{14.1} & \bl{30.7} & \bl{18.0} & \bl{32.3} & \bl{21.9} \\
RLAWM          & \textbf{57.6} & \textbf{39.6} & \textbf{23.9} & \textbf{64.0} & \textbf{47.9} & \textbf{26.4} & 21.9 & \textbf{28.0} & \textbf{42.7} & \textbf{20.5} & \textbf{38.9} & \textbf{29.0} \\
\midrule
\rowcolor{gray!18}\multicolumn{13}{@{}l}{\textit{Backbone: Qwen3-VL-4B-Instruct}} \\
\bl{Vanilla LLM}        & \bl{22.4} & \bl{19.6} & \bl{13.3} & \bl{52.7} & \bl{28.6} & \bl{10.3} & \bl{11.0} & \bl{ 9.0} & \bl{25.5} & \bl{ 3.9} & \bl{23.1} & \bl{13.1} \\
\bl{RLVR}               & \bl{27.1} & \bl{22.0} & \bl{14.2} & \bl{52.0} & \bl{30.4} & \bl{16.4} & \bl{17.1} & \bl{ 9.8} & \bl{23.0} & \bl{16.0} & \bl{30.9} & \bl{17.7} \\
\bl{Direct-Likert}      & \bl{31.4} & \bl{20.2} & \bl{ 9.7} & \bl{60.7} & \bl{32.6} & \bl{\textbf{20.4}} & \bl{ 8.5} & \bl{11.6} & \bl{\textbf{31.8}} & \bl{12.8} & \bl{27.8} & \bl{17.9} \\
\bl{Reference-Likert}   & \bl{29.8} & \bl{26.0} & \bl{15.8} & \bl{50.7} & \bl{32.0} & \bl{19.9} & \bl{13.6} & \bl{ 8.2} & \bl{28.1} & \bl{17.4} & \bl{23.3} & \bl{17.8} \\
\bl{Rubrics as Rewards}                & \bl{25.7} & \bl{23.1} & \bl{13.6} & \bl{46.0} & \bl{28.4} & \bl{17.3} & \bl{14.3} & \bl{ 9.8} & \bl{19.8} & \bl{19.4} & \bl{25.0} & \bl{16.7} \\
RLAWM          & \textbf{51.5} & \textbf{37.0} & \textbf{23.1} & \textbf{61.3} & \textbf{44.8} & 20.2 & \textbf{19.6} & \textbf{12.9} & 31.3 & \textbf{24.3} & \textbf{32.2} & \textbf{22.2} \\
\midrule
\rowcolor{gray!18}\multicolumn{13}{@{}l}{\textit{Backbone: Gemma-3-12B-IT}} \\
\bl{Vanilla LLM}        & \bl{11.4} & \bl{16.1} & \bl{11.1} & \bl{34.7} & \bl{19.3} & \bl{12.3} & \bl{ 3.5} & \bl{14.9} & \bl{25.1} & \bl{ 5.6} & \bl{22.2} & \bl{13.2} \\
\bl{RLVR}               & \bl{15.9} & \bl{17.1} & \bl{14.1} & \bl{38.7} & \bl{22.5} & \bl{13.6} & \bl{ 9.3} & \bl{18.1} & \bl{21.1} & \bl{ 9.5} & \bl{19.1} & \bl{14.8} \\
\bl{Direct-Likert}      & \bl{18.5} & \bl{15.5} & \bl{17.4} & \bl{37.4} & \bl{23.0} & \bl{18.3} & \bl{ 9.3} & \bl{13.6} & \bl{\textbf{26.0}} & \bl{11.3} & \bl{25.9} & \bl{16.7} \\
\bl{Reference-Likert}   & \bl{17.7} & \bl{14.2} & \bl{16.3} & \bl{34.7} & \bl{21.5} & \bl{15.2} & \bl{ 8.8} & \bl{\textbf{18.5}} & \bl{22.9} & \bl{12.4} & \bl{21.4} & \bl{16.1} \\
\bl{Rubrics as Rewards}                & \bl{15.6} & \bl{17.0} & \bl{13.9} & \bl{31.9} & \bl{20.4} & \bl{15.1} & \bl{ 7.6} & \bl{15.2} & \bl{22.6} & \bl{ 8.9} & \bl{21.7} & \bl{14.7} \\
RLAWM         & \textbf{35.6} & \textbf{19.4} & \textbf{17.8} & \textbf{44.9} & \textbf{30.6} & \textbf{19.7} & \textbf{11.3} & 15.4 & 24.5 & \textbf{14.9} & \textbf{30.6} & \textbf{18.3} \\
\bottomrule
\end{tabular}%
}
\end{table}

Table~\ref{tab:main_results} shows that RLAWM outperforms the strongest baselines across all backbones and datasets. On PhysReason, it gains up to $+12.2$ points, with consistent gains on PHYSICS-text\_only. Appendix~\ref{app:seeds} shows that seed-to-seed variation is far smaller than the margin over the strongest baseline.

The margin is largest on \textit{Easy} and \textit{Medium} problems (11--20 points) and narrows on \textit{Difficult} ones. This supports our hypothesis: when the model already has sufficient reasoning ability, modeling drift is the primary bottleneck rather than raw reasoning capacity, and optimizing axiomatic formulation unlocks that capacity. On \textit{Difficult} problems, the lower ceiling leaves less room for improvement. The margin thus tracks the reasoning ability masked by modeling drift.

Furthermore, performance improvements vary across PHYSICS subdomains. We attribute this to how explicitly different fields expose their physical structures to our checker; the PMF metric more readily captures the explicit coordinate frames of E\&M and Optics compared to the abstract operator algebra of Quantum and Statistical Mechanics.

\subsection{SubPass@k: Expanding the Reasoning Boundary}
\label{sec:pass_k}

\begin{figure}[!t]
  \centering
  \begin{minipage}[t]{0.4\linewidth}
    \centering
    \includegraphics[width=0.9\linewidth]{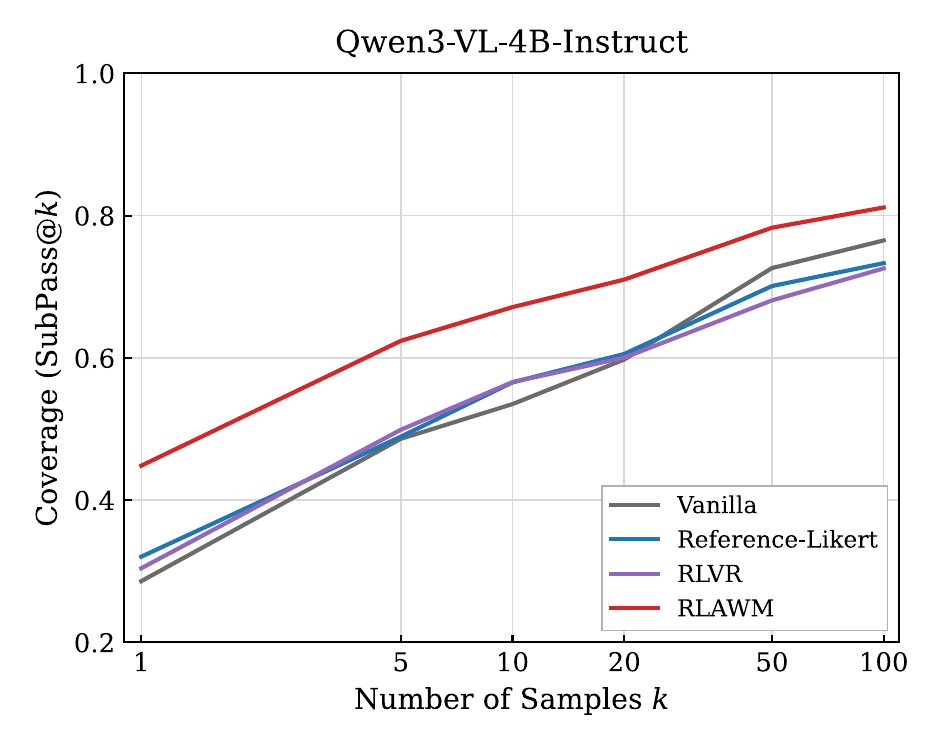}
    \subcaption{Qwen3-VL-4B-Instruct}
    \label{fig:passk_4b}
  \end{minipage}\hspace{0.04\linewidth}
  \begin{minipage}[t]{0.4\linewidth}
    \centering
    \includegraphics[width=0.9\linewidth]{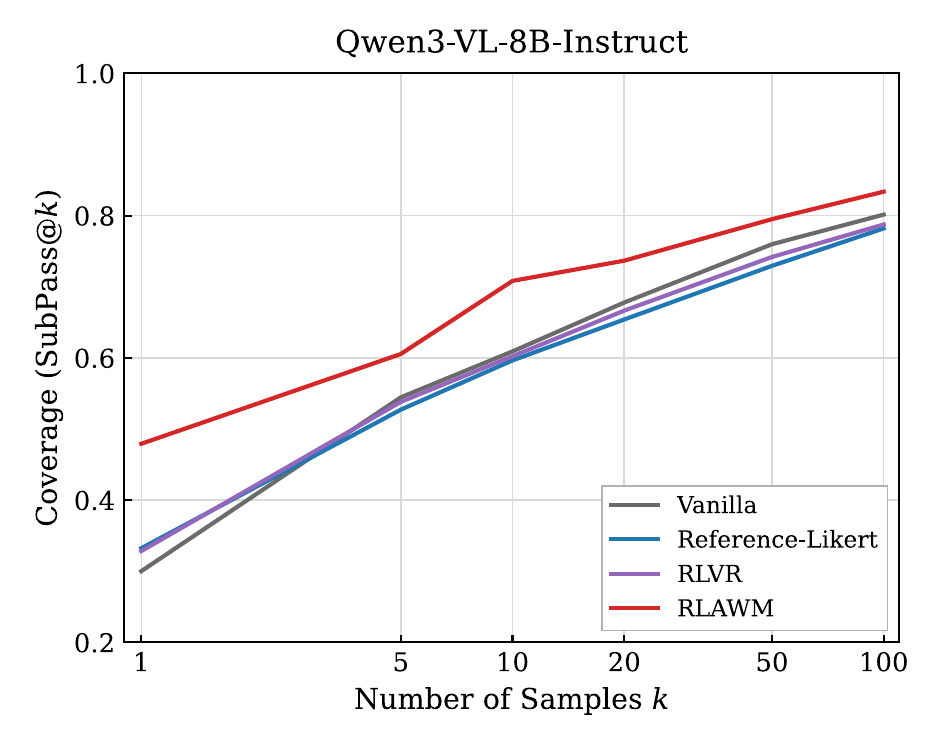}
    \subcaption{Qwen3-VL-8B-Instruct}
    \label{fig:passk_8b}
  \end{minipage}
  \caption{\textbf{SubPass@$k$ curves across $k\in\{1,5,10,20,50,100\}$ on PhysReason.}
    As $k$ grows, RLVR and Reference-Likert fall below the base model.
    RLAWM stays on top at every $k$.}
  \label{fig:passk}
\end{figure}

Figure~\ref{fig:passk} plots SubPass@$k$ on the Qwen3-VL backbones. As $k$ grows, RLVR and Reference-Likert fall below the base model. This is the diversity collapse observed by \mbox{\citet{yue2025rlvr_limit}}. In contrast, RLAWM improves monotonically, expanding the reasoning boundary.

This scaling behavior strongly validates our decoupled reward design. Outcome-only and text-judgment rewards often overfit to specific reasoning templates, penalizing valid alternative paths and compressing diversity. RLAWM avoids prescribing a unique trace: $R_M$ anchors the abstract axioms, while $R_S$ acts strictly as a filter for physical consistency. This allows the policy to freely explore diverse \texttt{representation} choices and calculation paths, steering it into a physically rigorous solution space without prematurely collapsing structural diversity.

\subsection{Out-of-Distribution Generalization}
\label{sec:generalization}

Table~\ref{tab:ood_olympiadbench}  shows that RLAWM (trained only on PhysReason) generalizes out of distribution along all three axes of \S\ref{sec:datasets}, with no checker involved at inference time.  On  the cross-difficulty OlympiadBench-Phys, RLAWM more than doubles the SubPass@1 performance, lifting it from RLVR's 6.2\% to 15.2\%. The gain holds across scales. On Qwen3-VL-4B, RLAWM raises OlympiadBench-Phys from 4.5\% (RLVR) to 11.3\%. On the cross-source PHYBench, RLAWM reaches 10.5\% against 8.7\% for the strongest baseline. Under the shared Zero-RL protocol, RLAWM leads RLVR by 10.0 points on the cross-domain AIME~2026 benchmark. \textbf{This confirms the OOD prediction of \S\ref{sec:intro}}: while text-based outcome rewards often overfit to in-distribution reasoning styles, systematically resolving modeling drift yields capabilities that transfer across all three axes.   Within physics, the gains come from two learned behaviors: less modeling drift and derivations that respect physical law. The cross-domain gain on AIME rests on the first behavior and on the shared axiomatic core of the two fields. Both behaviors are learned by the policy, so the gains transfer with no test-time support.

%% ============================================================
%% §8  Ablations and Additional Analyses
%% ============================================================
\section{Ablations and Additional Analyses}
\label{sec:ablations}

\subsection{Necessity of Each Reward Component}
\label{sec:ablation_per_stage}

\begin{table}[!t]
\centering
\caption{\textbf{Per-stage reward ablation.}
SubPass@1 (\%); best per column per backbone in \textbf{bold}.}
\label{tab:ablation_per_stage}
\scriptsize
\setlength{\tabcolsep}{3pt}
\renewcommand{\arraystretch}{1.08}
\resizebox{\linewidth}{!}{%
\begin{tabular}{@{}l ccccc ccccccc@{}}
\toprule
& \multicolumn{5}{c}{\textbf{PhysReason}} & \multicolumn{7}{c}{\textbf{PHYSICS-text\_only}} \\
\cmidrule(lr){2-6}\cmidrule(lr){7-13}
Reward & Easy & Medium & Difficult & Knowledge & Overall & Stat.~Mech. & Quantum & Atomic & E\&M & Mechanics & Optics & Overall \\
\midrule
\rowcolor{gray!18}\multicolumn{13}{@{}l}{\textit{Backbone: Qwen3-VL-8B-Instruct}} \\
$R_M{+}R_A$        & 48.3 & \textbf{39.8} & 17.6 & 63.0 & 44.0 & \textbf{28.2} & 19.7 & 18.2 & 33.5 & 20.0 & 36.1 & 25.1 \\
$R_S{+}R_A$        & 47.6 & 36.3 & \textbf{24.2} & \textbf{66.0} & 45.3 & 25.4 & 19.4 & 19.5 & 37.3 & 17.2 & \textbf{44.4} & 25.7 \\
\addlinespace[2pt]
Joint $R_M{+}R_S{+}R_A$\,$^{\dagger}$ & \textbf{57.6} & 39.6 & 23.9 & 64.0 & \textbf{47.9} & 26.4 & \textbf{21.9} & \textbf{28.0} & \textbf{42.7} & \textbf{20.5} & 38.9 & \textbf{29.0} \\
\midrule
\rowcolor{gray!18}\multicolumn{13}{@{}l}{\textit{Backbone: Qwen3-VL-4B-Instruct}} \\
$R_M{+}R_A$        & 40.2 & 36.0 & 19.8 & \textbf{68.0} & 43.0 & 18.9 & 12.5 & 11.2 & 30.1 & 15.6 & \textbf{33.3} & 18.9 \\
$R_S{+}R_A$        & 51.4 & 30.5 & 20.6 & 60.7 & 42.5 & 14.9 & 17.1 &  9.9 & \textbf{31.8} & 16.5 & 31.6 & 19.0 \\
\addlinespace[2pt]
Joint $R_M{+}R_S{+}R_A$\,$^{\dagger}$ & \textbf{51.5} & \textbf{37.0} & \textbf{23.1} & 61.3 & \textbf{44.8} & \textbf{20.2} & \textbf{19.6} & \textbf{12.9} & 31.3 & \textbf{24.3} & 32.2 & \textbf{22.2} \\
\bottomrule
\end{tabular}%
}
\end{table}

\begin{table}[!t]
\centering
\begin{minipage}[b]{0.47\linewidth}
\centering
\caption{\textbf{OOD generalization.} Zero-shot SubPass@1 (\%) of Qwen3-VL policies trained only on PhysReason dataset.}
\vspace{6pt}
\label{tab:ood_olympiadbench}
\scriptsize
\setlength{\tabcolsep}{3pt}
\renewcommand{\arraystretch}{1.0}
\begin{tabular}{@{}l c c c c@{}}
\toprule
& \multicolumn{2}{c}{Olympiad} & PHYBench & AIME \\
\cmidrule(lr){2-3}\cmidrule(lr){4-4}\cmidrule(l){5-5}
Method & 4B & 8B & 8B & 8B \\
\midrule
\bl{RLVR}             & \bl{4.5}      & \bl{6.2}      & \bl{5.8}      & \bl{46.7} \\
\bl{Reference-Likert} & \bl{3.1}      & \bl{4.4}      & \bl{8.7}      & \bl{43.3} \\
RLAWM                 & \textbf{11.3} & \textbf{15.2} & \textbf{10.5} & \textbf{56.7} \\
\bottomrule
\end{tabular}
\end{minipage}\hfill
\begin{minipage}[b]{0.47\linewidth}
\centering
\caption{\textbf{Reward sparsity with and without the checker.} All GRPO groups (8 rollouts each) from training on PhysReason dataset with Qwen3-VL-8B-Instruct model.}
\vspace{4pt}
\label{tab:sparsity}
\scriptsize
\setlength{\tabcolsep}{3pt}
\begin{tabular}{@{}l c c@{}}
\toprule
Metric & Outcome & + Checker \\
\midrule
Nonzero-variance groups & 70.1\% & 90.7\% \\
Zero-var.\ groups recovered & -- & 69.0\% \\
All-wrong, nonzero reward & 0\% & 64.1\% \\
Median reward & 0 & 0.378 \\
\bottomrule
\end{tabular}
\end{minipage}
\end{table}

\textbf{$R_M$ and $R_S$ target distinct, orthogonal failure modes.}
To test whether $R_M$ and $R_S$ each contribute beyond the answer reward $R_A$, we train each backbone with $R_M{+}R_A$, $R_S{+}R_A$, and the full joint $R_M{+}R_S{+}R_A$, holding all other settings fixed.
Both two-reward configurations in Table~\ref{tab:ablation_per_stage} exceed every non-RLAWM baseline in Table~\ref{tab:main_results}, confirming that both rewards reach the modeling-drift problem.
Once we split by difficulty and subdomain, however, the two configurations take separated winning regions.
$R_S{+}R_A$ takes a decisive lead on Difficult (8B 24.2 vs.\ 17.6; 4B 20.6 vs.\ 19.8), because long derivation chains make execution-level inconsistency the dominant failure.
$R_M{+}R_A$ instead leads on cells where the bottleneck is committing to the right governing law (4B Knowledge: 68.0 vs.\ 60.7; 8B PHYSICS Stat.\ Mech.: 28.2 vs.\ 25.4).
The joint reward therefore cannot be replaced by either two-reward configuration: on Easy, where both failure modes co-occur, joint produces the largest synergistic lift; on regime-extreme cells, joint stays close to the dominant configuration.

\subsection{What Training Signal Does the Solving Reward Add?}
\label{sec:signal_analysis}

\textbf{$R_S$ adds a dense and largely faithful training signal.}
GRPO learns only from reward differences within a group.
A group whose rollouts all score the same gives no gradient.
As detailed in Appendix~\ref{app:checker_diag}, we examine every GRPO group from RLAWM training on PhysReason dataset with Qwen3-VL-8B model.
As shown in Table~\ref{tab:sparsity}, 29.9\% of groups have zero reward variance under the outcome reward alone.
Adding the checker lowers this share to 9.3\%.
Each checker runs 7.34 assertions per rollout on average, so $R_S$ is graded rather than binary.
As a result, 64.1\% of all-wrong rollouts still receive a nonzero $R_S$, and hard problems stay in training.
The signal also rarely punishes valid work.
Only 4.9\% of correct, successfully executed rollouts are fully rejected.
False positives reach 15.8\%.
They arise when an error lies outside the checker's coverage, most often in a wrong final simplification.
Overall, $R_S$ supplies gradient across a wide range of cases where the outcome reward is silent, and $R_A$ covers what the checker cannot see.

\subsection{Sensitivity to the Modeling-Reward Scorer}
\label{sec:ablation_scorer}

\begin{table}[!t]
\centering
\caption{\textbf{Modeling-scorer ablation (PhysReason).}
SubPass@1 (\%); best per column within each backbone in \textbf{bold}.
\emph{4o}: \texttt{gpt-4o-mini}; \emph{qwen}: \texttt{Qwen3-VL-4B}.}

\label{tab:ablation_scorer}
\scriptsize
\setlength{\tabcolsep}{3pt}
\renewcommand{\arraystretch}{1.0}
\resizebox{\linewidth}{!}{%
\begin{tabular}{@{}l l ccccc ccccc@{}}
\toprule
& & \multicolumn{5}{c}{\textbf{Qwen3-VL-8B-Instruct}} & \multicolumn{5}{c}{\textbf{Qwen3-VL-4B-Instruct}} \\
\cmidrule(lr){3-7}\cmidrule(lr){8-12}
$R_M$  & $R_M$'s Scorer & Easy & Medium & Difficult & Knowledge & Overall & Easy & Medium & Difficult & Knowledge & Overall \\
\midrule
PMF    & 4o    & \textbf{57.6} & \textbf{39.6} & \textbf{23.9} & 64.0          & \textbf{47.9} & \textbf{51.5} & \textbf{37.0} & \textbf{23.1} & 61.3          & \textbf{44.8} \\
PMF    & qwen  & 55.1          & 31.6          & 20.5          & \textbf{68.0} & 45.8          & 44.8          & 36.6          & 17.4          & \textbf{65.3} & 43.0          \\
Judge  & 4o    & 57.3          & 36.5          & 21.3          & 66.0          & 47.1          & 50.5          & 34.2          & 19.7          & 64.0          & 44.0          \\
Judge  & qwen  & 51.4          & 30.9          & 19.8          & 66.7          & 44.2          & 43.8          & 31.1          & 16.1          & 62.0          & 40.2          \\
\bottomrule
\end{tabular}%
}
\end{table}

\textbf{PMF tolerates scorer weakening better than Judge.} 
As an alternative to PMF, we explore an LLM-as-Judge for $R_M$ that scores the modeling output from $1$ to $5$ (normalized to $[0,1]$; Appendix~\ref{app:judge_prompts}). To test how much each design relies on an external scorer, we swap the scorer from GPT-4o-mini to Qwen3-VL-4B (Table~\ref{tab:ablation_scorer}). While PMF outperforms Judge under GPT-4o-mini ($+0.8$ points on both backbones), the gap widens significantly (up to $+2.8$) under the weaker Qwen3-VL-4B scorer. Judge suffers roughly twice the performance drop of PMF on the 4B policy when the scorer degrades.

This resilience traces back to the critical path. PMF restricts the LLM to schema-bounded motif extraction and delegates scoring to a deterministic F1. The LLM only needs to parse text into tuples reliably—a skill that degrades gracefully. Judge, by contrast, requires the scorer to reason end-to-end about the physics, causing the weaker model's logic flaws to propagate directly into a noisy reward signal. Notably, even with a 4B scorer, RLAWM variants outperform every non-RLAWM baseline (Table~\ref{tab:main_results}), proving the pipeline remains highly effective without a massive external LLM.

\subsection{Disentangling Reward Signal from SFT Restructuring}
\label{sec:ablation_decouple}

\begin{table}[!t]
\centering
\caption{\textbf{Reward signal vs.\ SFT data (PhysReason, Qwen3-VL-8B).}
SubPass@1 (\%).
\emph{Zero-RL}: GRPO without SFT.
\emph{M/S SFT}: Reference-Likert trained on RLAWM's restructured SFT data.}

\label{tab:zero_rl}
\footnotesize
\setlength{\tabcolsep}{6pt}
\renewcommand{\arraystretch}{1.05}
\begin{tabular}{@{}l l ccccc@{}}
\toprule
Reward & Setting & Easy & Medium & Difficult & Knowledge & Overall \\
\midrule
Reference-Likert    & Zero-RL & 41.7          & 17.9          & 10.9          & 48.7          & 31.4 \\
Reference-Likert    & M/S SFT & 43.8          & 18.6          & 11.1          & 51.3          & 32.9 \\
RLAWM               & Zero-RL & \textbf{49.2} & \textbf{23.9} & \textbf{15.8} & \textbf{54.0} & \textbf{37.3} \\
\bottomrule
\end{tabular}
\end{table}

To isolate the contribution of our reward design from the schema-restructured SFT data, we run two further ablations on the Qwen3-VL-8B model. First, \textit{Zero-RL} applies GRPO directly to the base policy without any prior SFT. As shown in Table~\ref{tab:zero_rl}, Zero-RL RLAWM reaches 37.3 Overall on PhysReason, outperforming Reference-Likert (31.4) by $+5.9$ points and leading across all four subsets. Second, we test whether providing Reference-Likert with our structured \texttt{[modeling]} / \texttt{[solving]} SFT data closes this performance gap. This addition improves Easy and Knowledge subsets by $+2.1$ and $+2.6$ respectively, but Medium and Difficult scores barely change. This indicates that while structured SFT helps alleviate modeling drift on simpler problems by teaching the policy to recognize correct underlying physical axioms, solving the harder problems intrinsically relies on the execution-level feedback provided by RLAWM's carefully designed rewards.

%% ============================================================
%% §9  Conclusion and Limitations
%% ============================================================
\section{Conclusion}
\label{sec:conclusion}

We propose RLAWM, an RL framework that mitigates modeling drift in physics reasoning via two hierarchical rewards (modeling and solving). Together, they turn the verification-derivation asymmetry of axiomatic domains into a trainable signal. The gains hold across backbones,  seeds, and three axes of distribution shift. RLAWM extends to other axiomatic domain with cheap verification, such as formal mathematics, as the cross-domain gain on AIME~2026 suggests. Future work will replace static checkers with agentic probes and develop modeling rewards without external scorers.

\section*{Acknowledgments}
We acknowledge Delta at the National Center for Supercomputing Applications through allocation CIS251209 from the Advanced Cyberinfrastructure Coordination Ecosystem: Services \& Support (ACCESS) program, which is supported by U.S. National Science Foundation grants \#2138259, \#2138286, \#2138307, \#2137603, and \#2138296.

{
\small
\bibliographystyle{plainnat}
\bibliography{references}
}

%%%%%%%%%%%%%%%%%%%%%%%%%%%%%%%%%%%%%%%%%%%%%%%%%%%%%%%%%%%%

\appendix
% Long breakable listings leave irregular amounts of content on appendix pages.
% Keep unused space at the page bottom instead of stretching inter-block glue.
\raggedbottom

\section{Limitations}
\label{app:limitations}

RLAWM requires extra models during training. A frozen LLM scores the \texttt{[modeling]} block, and a frozen translator compiles the pseudocode. Another LLM synthesizes each checker from a fixed theorem library. These components raise training cost, and the static checkers never adapt to rollouts.

\section{Belief-State Framing and the Meaning of Axiomatic}
\label{app:belief}

\subsection{AWM is the Policy's Internal Belief, not a Separate World Model}

We use AWM to refer to the policy's internal belief about a problem's physical
structure during reasoning.
This belief is implicit in the LLM's parameters and hidden states.
The setup differs from work in which the world model is a separate, explicitly
trained predictor, as in the Dreamer line, Sora, and Genie.
There the world model is a dedicated network trained to predict the next frame or
the next state.
RLAWM introduces no new trainable network; the policy itself plays both the
reasoning role and the belief-carrying role.

The structured object $\hat{m}$ in \S\ref{sec:modeling_reward} is not the AWM itself
but its externalization.
The policy writes its implicit belief into the five-field \texttt{[modeling]} block
(\S\ref{sec:modeling_reward}), so the reward can read and score it.
Without externalization the reward cannot reach the belief.
This distinction has two practical implications.
First, RLAWM shapes the policy's belief only through reward, and adds no parameter
budget.
Second, the goal is not to make a world model more predictively accurate.
It is to align the physical world the policy adopts during reasoning with the
world the problem specifies, and to improve reasoning performance through this
alignment.

\subsection{What ``Axiomatic'' Means}

The true structure of a physics problem is defined by a set of axioms rather than
by a state trajectory.
The set includes conservation laws of energy, momentum, and charge; symmetries;
boundary conditions; regime assumptions such as a frictionless surface, a massless
string, an ideal gas, a small-angle approximation, or a non-relativistic limit; and
any approximations valid within the regime.
The AWM is the policy's internal representation of this axiom set, not a
predictive distribution over future states.

This contrasts with the role of a world model in game playing or robotic control.
In those settings, modeling correctly means predicting the next state accurately.
In axiomatic domains, modeling correctly means identifying the axiom set the
problem belongs to and reasoning consistently within it.
Modeling drift is therefore not a prediction error about future states.
It is the policy silently placing the problem into the wrong axiom set.

\section{Notation}
\label{app:notation}

We summarize the core notation used throughout the paper in Table~\ref{tab:notation}.
\begin{table}[!htbp]
  \centering
  \caption{Summary of core notation used throughout the paper.}
  \label{tab:notation}
  \small
  \begin{tabular}{@{}llp{0.55\linewidth}@{}}
    \toprule
    \textbf{Notation} & \textbf{Symbol} & \textbf{Description} \\
    \midrule
    \multicolumn{3}{@{}l}{\emph{Modeling reward (\S\ref{sec:modeling_reward})}} \\
    Policy                   & $\pi_\theta$       & LLM-parameterized policy \\
    Externalized AWM         & $\hat{m}$          & Policy's structured externalization of its belief (the five-field \texttt{[modeling]} block) \\
    Reference structure      & $m^*$              & An operational reference representation constructed from the problem statement and its ground-truth solution. It serves as a proxy for the problem's physical axiom set rather than as a separately formalized axiomatic theory. \\
    Per-motif F1             & $\mathrm{F1}_{\text{QEL/VCL/AEB}}$ & PMF motif scores \\
    Modeling reward          & $R_M$              & $R_M^{\text{PMF}}$ (PMF F1 over the externalized AWM) \\
    \midrule
    \multicolumn{3}{@{}l}{\emph{Solving reward (\S\ref{sec:solving_reward})}} \\
    Pseudocode               & $c$                & Policy's reasoning as pseudocode \\
    Translator               & $f$                & Frozen pseudocode-to-Python mapping (training only) \\
    AWM instance             & $\hat{w} = f(c)$   & Materialized executable instance \\
    Checker                  & $g$                &  Physics checker bound per problem from the theorem library \\
    Solving reward           & $R_S$              & Fraction of checker tests passed by $\hat{w}$ \\
    \midrule
    \multicolumn{3}{@{}l}{\emph{Joint optimization (\S\ref{sec:joint_objective})}} \\
    Final-answer reward      & $R_A$              & Correctness of the sub-answers in \texttt{[ans]} against the ground truth \\
    Joint weights            & $(\lambda_M, \lambda_S, \lambda_A)$  & $(0.4,\, 0.4,\, 0.2)$ \\
    Joint reward             & $R_{\text{joint}}$ & $\lambda_M R_M + \lambda_S R_S + \lambda_A R_A$ \\
    \bottomrule
  \end{tabular}
\end{table}
\FloatBarrier

\section{Dataset Details}
\label{app:datasets}

This appendix consolidates the dataset preparation details summarized in
\S\ref{sec:datasets}.
Table~\ref{tab:datasets} reports per-split sizes and composition for the  five
datasets used in this paper.

\begin{table}[!t]
\centering
\small
\caption{Dataset overview. We split PhysReason and PHYSICS-text\_only randomly
$9{:}1$ into training and test partitions; OlympiadBench-Phys, PHYBench, and
AIME~2026 are evaluation-only. PhysReason categories combine three reasoning-difficulty
tiers (\emph{easy} / \emph{medium} / \emph{difficult}) with one
knowledge-recall category (\emph{knowledge}); PHYSICS-text\_only subdomains are
taken directly from the source dataset and listed in test-split frequency
order.  PhysReason, PHYSICS, OlympiadBench, and PHYBench are released under the MIT License. AIME~2026 is taken from MathArena and released under CC BY-NC-SA 4.0.}
\label{tab:datasets}
\scriptsize
\setlength{\tabcolsep}{3pt}
\renewcommand{\arraystretch}{1.08}
\begin{tabularx}{\linewidth}{@{}>{\raggedright\arraybackslash}p{0.22\linewidth}clrY@{}}
\toprule
Dataset & License & Split & \# & Composition \\
\midrule
PhysReason~\citep{zhang2025physreasoncomprehensivebenchmarkphysicsbased}
                   & MIT & train & 1{,}080 & easy 24.5\% / medium 25.0\% / difficult 25.0\% / knowledge 25.5\% \\
                   &     & test & 120
                   & easy 25.0\% / medium 25.0\% / difficult 20.8\% / knowledge 29.2\% \\
\addlinespace
PHYSICS-text\_only~\citep{feng2025physicsbenchmarkingfoundationmodels}
                   & MIT & train & 897     & thermodynamics \& statistical mechanics 21.3\% / quantum mechanics 20.3\% / atomic physics 18.6\% / electromagnetism 17.1\% / mechanics 13.4\% / optics 9.4\% \\
                   &     & test  & 99      & thermodynamics \& statistical mechanics 21.2\% / quantum mechanics 20.2\% / atomic physics 19.2\% / electromagnetism 17.2\% / mechanics 13.1\% / optics 9.1\% \\
\addlinespace
OlympiadBench-Phys~\citep{he2024olympiadbenchchallengingbenchmarkpromoting}
                   & MIT & test  & 236     & text-only, English, physics, with verifiable ground truth (\texttt{OE\_TO\_physics\_en\_COMP}) \\
\addlinespace
PHYBench~\citep{PHYBENCH_NeurIPS2025}
                   & MIT & test  & 500     & full set; original text-only problems written by physics students, symbolic-expression answers, high-school to olympiad difficulty \\
\addlinespace
AIME~2026~\citep{dekoninck2026matharena}
                   & CC BY-NC-SA 4.0 & test  & 30     & full set (AIME I and II); competition mathematics, used for cross-domain evaluation of policies trained on PhysReason with Zero-RL \\
\bottomrule
\end{tabularx}
\end{table}
\FloatBarrier

\subsubsection*{Dataset case: PHYSICS-text\_only}
\begin{tcblisting}{
  samplebox,
  breakable,
  title={Representative Problem from PHYSICS-text\_only},
}
{
  "id": "atomic/1-13",
  "questions": "The Thomas-Fermi model of atoms describes the electron cloud in an atom as a continuous distribution rho(x) of charge. An individual electron is assumed to move in the potential determined by the nucleus of charge Ze and of this cloud. Derive the equation for the electrostatic potential in the following stages. (a) By assuming the charge cloud adjusts itself locally until the electrons at Fermi sphere have zero energy, find a relation between the potential phi and the Fermi momentum p_F. (b) Use the relation derived in (a) to obtain an algebraic relation between the charge density rho(x) and the potential phi(x). (c) Insert the result of (b) in Poisson's equation to obtain a nonlinear partial differential equation for phi.",
  "final_answers": [
    "p_f^2 = 2me phi(x)",
    "rho(x) = e/(3 pi^2 hbar^3) [2me phi(x)]^(3/2)",
    "(1/r) d^2/dr^2 [r phi(r)] = 4e/(3 pi hbar^3) [2me phi(r)]^(3/2)"
  ]
}
\end{tcblisting}

\subsubsection*{Dataset case: PhysReason}
\begin{tcolorbox}[
  samplebox,
  breakable,
  listing only=false,
  title={Representative Problem from PhysReason},
]
  \centering
  \includegraphics[width=0.78\linewidth]{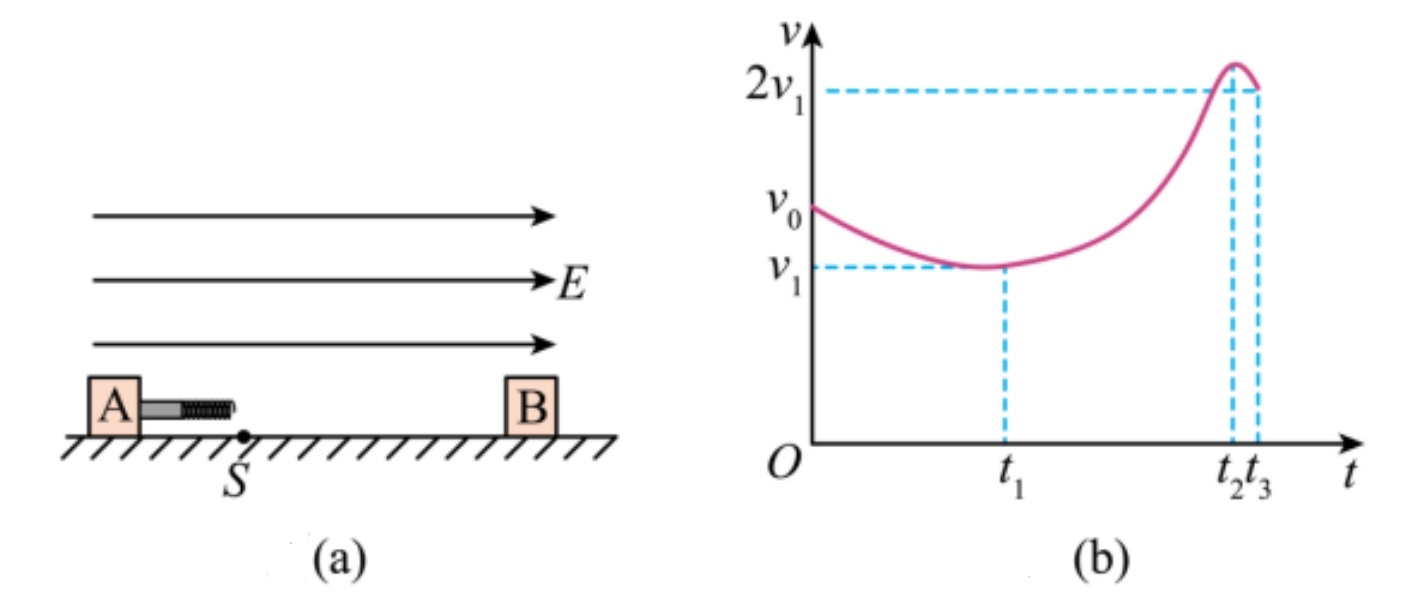}
  \par\vspace{0.75em}
\begin{lstlisting}[style=appendixlistingtiny]
{
  "context": "As shown in figure (a), there are two small sliders A and B, each with mass m, on a rough, insulated horizontal plane. Their charges are q (q>0) and -q. A light, smooth, insulated rod and a light, insulated spring are fixed at the right end of A, with the spring initially at natural length. A uniform electric field of magnitude E is directed horizontally to the right. The maximum static friction and sliding friction on A and B both have magnitude 2qE. At t=0, A moves right with initial velocity v0, while B is at rest. At time t1, A reaches position S with velocity v1, and the spring is not yet in contact with B. At time t2, the velocity of A reaches its maximum, and the spring force magnitude is 3qE. Immediately before the rod collides with B, A has velocity 2v1 at t=t3. The v-t graph of A over 0 to t3 is shown in figure (b), where v1 is the minimum speed and t1, t2, t3 are unknown. During the motion, A and B remain on the same straight line, charges remain constant, Coulomb force is modeled as point-charge electrostatic force in vacuum with constant k, there is no mechanical energy loss when B contacts the spring, and the spring remains elastic.",
  "sub_question_1": "Determine the work done by the net external force on A during 0 to t1.",
  "sub_question_2": "Determine the distance between A and B at time t1.",
  "sub_question_3": "Calculate the total work done by the uniform electric field on A and B during t1 to t2.",
  "sub_question_4": "If the initial velocity of A is increased such that its velocity at position S is 2v1, determine the velocity of A immediately before the rod collides with B.",
  "answer": [
    "1/2 m v1^2 - 1/2 m v0^2",
    "sqrt(k q / E)",
    "q sqrt(k q E) / 2",
    "(1 + sqrt(3)) v1"
  ]
}
\end{lstlisting}
\end{tcolorbox}

\section{Model Assets and Licenses}
\label{app:assets}

Table~\ref{tab:assets} lists every pre-trained model used in this paper, together with its role in the pipeline and license. Open-weight backbones and the frozen translator are obtained from their public Hugging Face releases; \texttt{gpt-4o-mini} is accessed exclusively through the OpenAI API. All assets are used in accordance with their respective licenses and terms of service.

\begin{table}[!ht]
\centering
\caption{Pre-trained model assets used in this paper.}
\label{tab:assets}
\scriptsize
\setlength{\tabcolsep}{4pt}
\renewcommand{\arraystretch}{1.1}
\begin{tabularx}{\linewidth}{@{}>{\raggedright\arraybackslash}p{0.30\linewidth}Yl@{}}
\toprule
Model & Role & License \\
\midrule
Qwen3-VL-8B-Instruct~\citep{bai2025qwen3vltechnicalreport}
  & Policy backbone (main and ablations) & Apache 2.0 \\
Qwen3-VL-4B-Instruct~\citep{bai2025qwen3vltechnicalreport}
  & Policy backbone; compact PMF scorer (ablation) & Apache 2.0 \\
Gemma-3-12B-IT~\citep{gemmateam2025gemma3technicalreport}
  & Policy backbone (cross-family verification) & Gemma Terms of Use \\
Qwen2.5-Coder-1.5B-Instruct~\citep{hui2024qwen25codertechnicalreport}
  & Frozen translator $f$ (pseudocode $\to$ Python) & Apache 2.0 \\
gpt-4o-mini-2024-07-18~\citep{openai2024gpt4technicalreport}
  & PMF scorer for $R_M$; Judge baseline & Proprietary (OpenAI API ToS) \\
\bottomrule
\end{tabularx}
\end{table}
\section{Case Study: Why We Need Rewards at Different Hierarchical Levels}
\label{app:case_studies}

This case study illustrates how the solving reward can complement
the modeling reward: stating valid physical premises does not ensure
that those premises are correctly applied during a derivation.
We compare two rollouts from the same training problem under the same
policy, presented below as condensed, schematic summaries.
Both state the relevant physical premises, but organize them using
different \texttt{representation} choices.
The first rollout correctly accounts for the work done along the arc.
The second omits a nonzero work contribution during solving, despite
identifying the relevant forces during modeling.
A work--energy invariant check can detect this execution-level error.

\subsubsection*{Problem}

\begin{tcolorbox}[
  appendixbox,
  title={Problem (PhysReason, multi-stage motion with electric field)},
]
A smooth horizontal surface $AB$ and a smooth quarter-circular arc guide
rail in the vertical plane are smoothly connected at point $B$.
The radius of the guide rail is $R$.
A positively charged small ball of mass $m$ is released from rest
after compressing a light spring to point $A$.
After leaving the spring, the ball's velocity at $B$ has magnitude
$\sqrt{gR}$.
The ball then moves along the track $BO$.
A rectangular coordinate system $xOy$ is established with $O$ as the
origin.
The positive $x$-axis points rightward and the positive $y$-axis
points upward. In this geometry, $B=(-R,-R)$ and $O=(0,0)$, so the
displacement from $B$ to $O$ is $(R,R)$.
The tangent to the track at $O$ is vertical, and the ball leaves the
track moving upward.
In the region $x \geq -R$, there is a uniform electric field at angle
$\theta = 45^\circ$ to the $x$-axis.
After entering the field, the magnitude of the electric force on the
ball is $\sqrt{2}\,mg$.
The charge of the ball remains constant; the gravitational acceleration
is $g$.

\medskip
\noindent\textbf{Sub-questions.}
\textbf{(Q1)} elastic potential energy of the spring at maximum
compression at $A$;
\textbf{(Q2)} magnitude of the ball's velocity at $O$;
\textbf{(Q3)} trajectory equation of the ball after passing $O$.
\end{tcolorbox}

\subsubsection*{Case 1: \texttt{timeline} construal $\to$ high reward}

\begin{tcblisting}{
  longpromptbox,
  colframe=green!50!black,
  title={Rollout with \texttt{timeline} representation (high reward)},
  listing options={
    style=appendixlistingtiny,
    escapeinside={(*}{*)},
  },
}
[modeling]
[objects]
...
[constraints]
...
[representation]
type: timeline

S1: A -> B
- Spring elastic potential energy becomes kinetic energy at B.

S2: B -> O
- The ball moves along the constrained quarter arc.
- Normal force does no work.
- Gravity and electric force both do work.
- Apply work-energy theorem.

S3: after O
- The ball leaves the rail with velocity v_O.
- Motion is under constant gravity and electric force.
- Apply force decomposition and kinematics.

[mapping]
...
[plan]
...

[solving]
...
[sub2]
[steps]
- Apply work-energy theorem from B to O:
(*\textcolor{green!50!black}{\quad W\_g + W\_E = (1/2) m v\_O\textasciicircum 2 - (1/2) m v\_B\textasciicircum 2.}*)
- W_g = -m g R; W_E = 2 m g R (the electric force has a horizontal projection
  q E cos 45 = m g acting over the horizontal displacement R, plus a vertical
  projection q E sin 45 = m g acting over the vertical rise R).
- Substitute v_B^2 = g R:
  m g R = (1/2) m v_O^2 - (1/2) m g R.
- Solve: (*\textcolor{green!50!black}{$v_O = \sqrt{3gR}$}*).

[sub3]
[steps]
- After O the net horizontal force is q E cos 45 = m g, so a_x = g.
- The vertical electric component q E sin 45 = m g cancels gravity, so a_y = 0.
- x = (1/2) g t^2, y = v_O t.
- Substitute v_O^2 = 3 g R:  y^2 = 6 R x.

[ans]
$E_p = \tfrac{1}{2} m g R$, \quad $v_O = \sqrt{3 g R}$, \quad $y^2 = 6 R x$.
\end{tcblisting}

\subsubsection*{Case 2: \texttt{interaction\_graph} construal $\to$ low reward}

\begin{tcblisting}{
  longpromptbox,
  colframe=red!60!black,
  title={Rollout with \texttt{interaction\_graph} representation (low reward)},
  listing options={
    style=appendixlistingtiny,
    escapeinside={(*}{*)},
  },
}
[modeling]
[objects]
...
[constraints]
...
[representation]
type: interaction_graph

Nodes:
- Small ball
- Spring
- Gravity
- Electric field
- Guide rail

Edges:
- Spring -> ball:
  transfers elastic potential energy before B.
- Gravity -> ball:
  downward force m g.
- Electric field -> ball:
  q E = sqrt(2) m g at 45 degrees.
- Guide rail -> ball:
  normal force constrains motion but does no work.

Interaction consequences:
- F_Ex = q E cos 45 = m g.
- F_Ey = q E sin 45 = m g.
- After O, F_Ey cancels gravity.
- After O, a_x = g and a_y = 0.

[mapping]
...
[plan]
...

[solving]
...
[sub2]
[steps]
- From the interaction graph, F_Ey = m g and gravity = m g downward.
- Therefore the vertical forces cancel.
- The normal force does no work.
(*\textcolor{red!70!black}{- The kinetic energy from B to O is unchanged.}*)
(*\textcolor{red!70!black}{- So $v_O = v_B = \sqrt{gR}$.}*)

[sub3]
[steps]
- After O: a_x = g, a_y = 0.
- x = (1/2) g t^2, y = v_O t.
- Substitute v_O^2 = g R:  y^2 = 2 R x.

[ans]
$E_p = \tfrac{1}{2} m g R$, \quad $v_O = \sqrt{g R}$ \emph{(low reward)}, \quad
$y^2 = 2 R x$ \emph{(low reward)}.
\end{tcblisting}

\paragraph{Why both rewards are needed.}

The relevant modeling commitments alone do not distinguish the
two solving outcomes. Both rollouts identify gravity, the electric
force, and the non-working normal force. The first explicitly accounts
for their work along the arc, whereas the second incorrectly infers
unchanged kinetic energy from force cancellation.

Gravity and the vertical electric force cancel throughout the
field region, including along $BO$. This cancellation does not eliminate
the horizontal electric force or its work.
The normal force can still affect the acceleration along the curved
track, but it does no work.
Consequently, the net work from $B$ to $O$ is
$W_g + W_E = -mgR + 2mgR = mgR$,
equivalently the work of the horizontal electric component,
$F_{E,x}\Delta x = mgR$.
The second rollout omits this contribution.

An invariant check compares the candidate's kinetic-energy
change, $\tfrac{1}{2}m(\hat{v}_O^2-v_B^2)$, with $mgR$.
For the second rollout, $\hat{v}_O=v_B$, so the reported change is zero
and the check fails.
Its subsequent trajectory $y^2=2Rx$ is internally consistent with
that incorrect velocity, illustrating why consistency among emitted
quantities alone is insufficient: those quantities must also satisfy
the physical relations implied by the problem.

This example illustrates the additional role of $R_S$ when
valid modeling commitments are incorrectly applied during solving.
$R_M$ evaluates the declared physical setup, while $R_S$ checks
selected physical relations in its executable instantiation.
The reward ablations in \S\ref{sec:ablation_per_stage} provide
quantitative comparisons of the individual reward combinations.

\section{Hyperparameters}
\label{app:hyperparams}

All methods in our comparison use the same optimization, batching, precision,
and decoding hyperparameters.  RLAWM and the baselines differ only in the reward
source used during GRPO, as described in \S\ref{sec:baselines}.
Table~\ref{tab:sft_hyperparams} lists the supervised fine-tuning setup, and
Table~\ref{tab:grpo_hyperparams} lists the GRPO setup.

\begin{table}[h]
\centering
\small
\caption{Main SFT hyperparameters shared by all methods.}
\label{tab:sft_hyperparams}
\begin{tabular}{@{}ll@{}}
\toprule
Hyperparameter & Value \\
\midrule
Fine-tuning type & Full fine-tuning \\
Vision tower & Frozen \\
Multimodal projector & Trainable \\
Language model & Trainable \\
Cutoff length & 8192 \\
Maximum samples & 5000 \\
Per-device batch size & 2 \\
Gradient accumulation steps & 8 \\
Effective batch size per device group & 16 \\
Learning rate & $1.0{\times}10^{-5}$ \\
Epochs & 4 \\
LR scheduler & Cosine \\
Warmup ratio & 0.1 \\
Random seed & 42, with a seed sweep in Appendix~\ref{app:seeds} \\
\bottomrule
\end{tabular}
\end{table}

\begin{table}[h]
\centering
\small
\caption{Main GRPO hyperparameters shared by RLAWM and all RL baselines.}
\label{tab:grpo_hyperparams}
\begin{tabular}{@{}ll@{}}
\toprule
Hyperparameter & Value \\
\midrule
Advantage estimator & GRPO \\
Normalize GRPO advantages by std. & Yes \\
KL in reward & No \\
Actor KL loss & Enabled \\
KL coefficient & 0.001 \\
KL loss type & Low-variance KL \\
Entropy coefficient & 0.01 \\
PPO clip ratio & 0.2 \\
Actor learning rate & $1.0{\times}10^{-6}$ \\
Train batch size & 24 \\
PPO mini-batch size & 24 \\
PPO micro-batch size per GPU & 2 \\
PPO epochs per update & 2 \\
Rollouts per prompt & 8 \\
Maximum prompt length & 4096 \\
Maximum response length & 8192 \\
Rollout maximum model length & 12288 \\
Maximum batched tokens & 65536 \\
Vision tower & Frozen \\
Precision & bfloat16 \\
Training epochs & 3 \\
Random seed & 42, with a seed sweep in Appendix~\ref{app:seeds} \\
\bottomrule
\end{tabular}
\end{table}

\paragraph{Compute.}
Each RL training run uses 4$\times$NVIDIA B200 GPUs: 3 for GRPO policy
training and 1 dedicated to a vLLM server hosting the frozen translator $f$
(\S\ref{sec:solving_reward}) that maps pseudocode rollouts into executable
Python for the checker. A full training run completes in roughly 6 hours
under this configuration.

\section{Robustness to Random Seeds}
\label{app:seeds}

RLAWM's gains are robust to the random seed. We retrain RLAWM end to end on
Qwen3-VL-8B-Instruct under three seeds. Every other setting is identical to the
main experiments in \S\ref{sec:eval_protocol} and Appendix~\ref{app:hyperparams}.
Table~\ref{tab:seeds} reports PhysReason SubPass@1 Overall for each run. The
three runs give $48.1 \pm 0.9$, so the seed-to-seed spread is below one point.
The worst run still exceeds the strongest baseline, Direct-Likert, by 10.2
points. The improvements reported in Table~\ref{tab:main_results} therefore
reflect the reward design rather than a favorable seed.

\begin{table}[h]
\centering
\small
\caption{\textbf{Seed robustness of RLAWM on PhysReason.} SubPass@1 Overall (\%) on Qwen3-VL-8B-Instruct; seed 42 is the run reported in Table~\ref{tab:main_results}.}
\label{tab:seeds}
\begin{tabular}{@{}l c@{}}
\toprule
Run & Overall \\
\midrule
\bl{Direct-Likert (strongest baseline)} & \bl{37.1} \\
RLAWM, seed 10  & 49.1 \\
RLAWM, seed 42 (reported)  & 47.9 \\
RLAWM, seed 100 & 47.3 \\
\midrule
Mean $\pm$ std & $48.1 \pm 0.9$ \\
\bottomrule
\end{tabular}
\end{table}

\section{Modeling Schema and SFT Data Synthesis}
\label{app:modeling_schema}

The SFT phase installs the structured \texttt{[modeling]} / \texttt{[solving]} output
format defined in \S\ref{sec:modeling_reward} and \S\ref{sec:solving_reward}.
We synthesize the SFT corpus by prompting GPT-4o-mini twice per problem, conditioning
on the dataset's ground-truth solution and target final answers.
The first call uses the modeling instruction below to elicit a JSON object
(a parsing-friendly intermediate format) with the five \texttt{[modeling]}
fields: \texttt{objects}, \texttt{constraints}, \texttt{representation},
\texttt{mapping}, and \texttt{plan}. The modeling reward scores all five.
\texttt{mapping} and \texttt{plan} also bridge modeling and solving
(\S\ref{sec:solving_reward}).
The second call uses the pseudocode-template instruction, populated per
problem with the expected sub-question tags, target answers, and a required
structural skeleton, to elicit a JSON object describing \texttt{givens},
\texttt{shared\_variables}, \texttt{global\_equations}, and an array of
\texttt{subproblems}.
Both calls share a single user template that wraps the problem context and
the list of sub-questions.
A post-processing script parses the two JSON outputs and assembles them into
the \texttt{[modeling]} and \texttt{[solving]} text blocks that form the SFT
target the policy is trained to emit.

\paragraph{Modeling-block instruction.}
\begin{tcblisting}{
  promptbox,
  title={Modeling-block instruction (sent to GPT-4o-mini)},
}
Generate the physics modeling phase as structured JSON.

### REQUIRED JSON FORMAT
You MUST output exactly these 5 keys:
{
  "objects": "...",
  "constraints": "...",
  "representation": "...",
  "mapping": "...",
  "plan": "..."
}

### FIELD GUIDE

#### "objects"
- List every physical object, entity, agent, and relevant environment component.
- For each object, include the key physical properties and the main forces or fields acting on it.
- Use concise bullet points only.
- Use at most 6 bullets.

#### "constraints"
- List explicit and implicit conditions.
- Include boundary conditions, conservation laws, sign conventions, admissible branches, positivity/range constraints, and any physical-validity conditions needed later in [solving].
- Use concise bullet points only.
- Use at most 10 bullets.

#### "representation"
- Choose exactly one of: coordinate, timeline, interaction_graph, state_table, or none.

#### "mapping"
- Convert the text constraints into governing equations or named physics principles.
- Format each line as: "[physics law/principle]" -> $formula$
- Include variable definitions and branch choices when they matter downstream.
- Write equations in stable symbolic form that can be reused directly in [solving].
- Use at most 6 lines.

#### "plan"
- One line per sub-question.
- Format exactly: Q{i}: ...
- Each line should say what quantity or condition is solved for, and the main law/relation used.
- If a later sub-question depends on an earlier intermediate result, say so explicitly.
- Keep each line short.

### BREVITY BUDGET
- Keep the entire JSON compact.
- Prefer short symbolic equations over explanatory prose.
- Do not repeat the same variable definitions across fields.
- The student-facing content should usually fit within about 1600 characters total.

### IMPORTANT
- Do NOT output [modeling], [solving], [pseudocode], [subN], [subans], or [ans].
- Do NOT wrap the answer in markdown fences.
- Optimize the modeling for downstream executable-style pseudocode: later [solving] should be able to reuse your variables, equations, and branch conditions directly.
\end{tcblisting}

\paragraph{Solving-block instruction.}
The placeholders \texttt{\{target\_answers\}}, \texttt{\{expected\_sub\_tags\}}, and
\texttt{\{required\_template\}} are filled per problem before the call.
\begin{tcblisting}{
  promptbox,
  title={Pseudocode instruction template (sent to GPT-4o-mini)},
}
Now write the pseudocode-style [solving] stage for all sub-questions.

### REQUIRED JSON FORMAT
{
  "pseudocode_plan": "...",
  "givens": ["..."],
  "shared_variables": ["..."],
  "global_equations": ["..."],
  "subproblems": [
    {
      "index": 1,
      "input_mapping": ["..."],
      "variables": ["..."],
      "equations": ["..."],
      "constraints": ["..."],
      "steps": ["..."],
      "emit": ["results['x'] = x"],
      "subans": "..."
    }
  ]
}

### TARGET FINAL ANSWERS
Use these answers as alignment hints only, while keeping the pseudocode self-consistent and reasoning-first.
Do not merely restate the answers. Equations, constraints, emitted quantities, and subans must support them.

[TARGET_ANSWERS]
{target_answers}

### PSEUDOCODE REQUIREMENTS
- Return structured fields, not tagged text. The script will assemble the final [solving] block for you.
- "givens", "shared_variables", and "global_equations" are optional arrays of short strings.
- "subproblems" must be a list whose entries correspond exactly to {expected_sub_tags}.
- Every subproblem object MUST include:
  - "index"
  - "emit"
  - "subans"
  - at least one non-empty field among "input_mapping", "equations", or "steps"
- Use natural-language pseudocode and symbolic equations. Do NOT write Python code.
- Reuse the [modeling] model. Do not invent a new model in [solving].
- "equations" and "constraints" are the primary execution specification. "steps" should only explain solve order and short local substitutions.
- "emit" must expose concrete computed quantities or symbolic threshold quantities that can support verification.
- NEVER emit undefined labels or string placeholders such as results['L_condition'] = L_condition.
- For inequalities or condition questions, emit the concrete boundary quantity (for example L_min, theta_max, or E_threshold), and let "subans" state the inequality or condition.
- "subans" must contain only the final answer for that sub-question, short and human-readable.
- The script will build the final [ans] block directly from the ordered "subans" values. Do not add a separate final-answer summary outside the JSON schema.
- Use ASCII-friendly variable names inside pseudocode content: theta, mu, delta_P, dt, dx, etc.
- Reuse the [modeling] model. Treat the target answers only as alignment hints, not as a substitute for equations or emitted quantities.

### REQUIRED STRUCTURAL SKELETON
{required_template}
\end{tcblisting}

\section{PMF Motif Taxonomy and Error-Mode Diagnosis}
\label{app:motif_taxonomy}

Physics Modeling Fidelity (PMF) decomposes a structured modeling output into
three motif sets.  Each motif type captures a different failure mode of the
externalized AWM: whether the right law is selected, whether equations are
coupled through shared quantities, and whether the assumptions required by each
equation are present.
For a policy output $\hat{m}$ and a reference model $m^*$, we extract the same
three motif sets from both sides and compute an F1 score for each type.

\paragraph{Quantity-Equation-Law (QEL).}
QEL motifs test whether the model chooses the correct physical principle and
whether the equation contains the required physical quantity.
Each motif has type
\[
  \texttt{(quantity, law, context)} .
\]
Here \texttt{quantity} is the physical variable being solved for or constrained,
\texttt{law} is the named principle or governing equation, and \texttt{context}
identifies the local physical scene in which the relation is used.
QEL therefore detects mistakes such as selecting the wrong theorem, omitting a
quantity from an otherwise plausible equation, or assigning a law to the wrong
stage of a multi-step problem.

\paragraph{Variable Coupling Link (VCL).}
VCL motifs test whether multiple equations are connected through the right
shared variable.  This is especially important in long-chain reasoning, where
the output of one relation becomes the input of another.
Each motif has type
\[
  \texttt{((law\_a, ctx\_a), shared\_var, (law\_b, ctx\_b))} .
\]
The motif records that \texttt{law\_a} in context \texttt{ctx\_a} and
\texttt{law\_b} in context \texttt{ctx\_b} must communicate through
\texttt{shared\_var}.  VCL therefore detects broken equation chains, inconsistent
intermediate variables, and cases where correct local equations fail to work
together globally.

\paragraph{Assumption-Equation Binding (AEB).}
AEB motifs test whether the conditions required for an equation to be valid are
present in the model.  A model can score well on QEL by choosing the right law
and variables, while still applying that law under invalid assumptions.
Each motif has type
\[
  \texttt{(assumption, law, context)} .
\]
Here \texttt{assumption} is a simplifying condition or boundary condition, such
as frictionless motion, constant volume, small-angle approximation, admissible
branch, or a stage-specific force balance.  AEB therefore detects missing
preconditions even when the equation text itself looks correct.

\begin{table}[h]
\centering
\caption{Diagnostic effect of common modeling errors on the three PMF motif
components. ``May decrease'' means that the component is affected only when the
local error changes the corresponding motif type.}
\label{tab:pmf_error_diagnosis}
\begin{tabular}{llll}
\toprule
Error type & QEL & VCL & AEB \\
\midrule
Wrong theorem selected & Decreases & May decrease & Unchanged \\
Correct theorem but missing variable & Decreases & May decrease & Unchanged \\
Broken interaction between equations & Unchanged & Decreases & Unchanged \\
Missed premise or assumption & Unchanged & Unchanged & Decreases \\
Missed object or physical stage & Decreases & Decreases & May decrease \\
\bottomrule
\end{tabular}
\end{table}

The three-way decomposition makes PMF diagnostic rather than merely scalar.
For example, if QEL drops while AEB is stable, the policy likely selected a wrong
law or omitted a key variable rather than forgetting a precondition.  If VCL drops
while QEL is stable, the model may have written locally correct equations but
failed to couple them through the right intermediate.  If AEB alone drops, the
policy has usually applied a formula outside its valid regime.

\section{PMF Extraction Prompt Templates}
\label{app:pmf_prompts}

PMF uses the same extraction prompt for the policy modeling output and the
reference modeling output.  The extractor maps a reasoning text into three motif
sets, then PMF computes motif-level F1 between the two extracted structures.

\paragraph{PMF motif-extraction system prompt.}
\begin{tcblisting}{
  promptbox,
  title={System prompt for PMF motif extraction},
}
You are a physics modelling analyst. Your task is to extract modelling facts from a physics reasoning text.

Extract three types of modelling tuples:

1. **QEL** (Quantity-Equation-Law): A physical quantity that appears in an equation, where that equation is based on a specific physics law applied in a specific physical context/stage/object.

2. **VCL** (Variable Coupling Link): A shared variable that connects two equations -- one equation outputs this variable, and another equation takes it as input. Only extract VCL when there is a clear multi-equation coupling.

3. **AEB** (Assumption-Equation Binding): A simplifying assumption that constrains the validity of a specific equation.

## Normalization rules
- All string values must use lowercase with underscores (e.g. "newton_second_law", "block_a", "incline_descent")
- `law` must be a named physics principle (e.g. "energy_conservation", "newton_second_law", "work_energy_theorem", "momentum_conservation", "coulombs_law", "ohms_law", "kinematics")
- `context` = the physical stage, object, or region where the equation applies (e.g. "incline_descent", "horizontal_surface", "block_a", "block_b", "collision", "free_fall")
- `quantity` / `shared_var` = symbolic name of the physical quantity (e.g. "v", "m", "g", "mu", "h", "theta", "t")
- `assumption` = a named simplifying condition (e.g. "frictionless_incline", "lossless_transition", "negligible_air_resistance", "point_mass", "uniform_field")

## Output format
You MUST output valid JSON with exactly these three keys. Empty arrays are allowed when no tuples of that type are present.

```json
{
  "qel": [
    {"quantity": "m", "law": "energy_conservation", "context": "incline_descent"}
  ],
  "vcl": [
    {"law_a": "energy_conservation", "ctx_a": "incline_descent",
     "shared_var": "v", "law_b": "work_energy_theorem", "ctx_b": "horizontal_surface"}
  ],
  "aeb": [
    {"assumption": "frictionless_incline", "law": "energy_conservation",
     "context": "incline_descent"}
  ]
}
```

Only extract motifs that are **explicitly stated or clearly implied** by the text. Do not invent motifs that are absent. If a tuple type has no instances, return an empty array for that key.
\end{tcblisting}

\section{LLM-Judge Prompt Templates}
\label{app:judge_prompts}

Like PMF, the LLM-as-Judge method takes the policy's full \texttt{[modeling]}
block as input. The judge is instructed to be conservative. It evaluates four
fields separately: \texttt{objects}, \texttt{constraints}, \texttt{representation},
and \texttt{mapping}. It returns a single integer score on a 1--5 scale with a
short justification. The system prompt used for this baseline is shown below.

\paragraph{Modeling-quality judge system prompt.}
\begin{tcblisting}{
  promptbox,
  title={LLM-as-Judge prompt for modeling quality},
}
You are a STRICT physics grading assistant who evaluates the quality of a student's physical-model setup -- the [modeling] section that precedes [solving]. You are critical by default; most real student attempts have noticeable flaws, so scores of 5 should be RARE and reserved for truly exceptional work.

A high-quality [modeling] section has four sub-sections:
  [objects]         -- every physical object, participant, field, medium, boundary, circuit element, environment component, with its initial state, attributes, and the forces/fields acting on it.
  [constraints]     -- every explicit AND implicit condition extracted from the problem, each paired with the physical law it invokes (e.g. Newton II, Kirchhoff, energy conservation, continuity, Gauss law).
  [representation]  -- a structured choice from Coordinate / Timeline / Interaction Graph / State Table, with enough detail to be machine-readable (axes, phases, nodes, cells).
  [mapping]         -- translation of every natural-language description into a math expression or constraint (e.g. "smooth" -> mu=0; "just leaves" -> N=0; "normally lit" -> U=U_rated & I=I_rated). A mapping that only covers the obvious givens is INSUFFICIENT.

Scoring rubric -- apply strictly, do not anchor on 5:
  5 (exceptional, rare):   All four sub-sections present, each correct AND complete. Every implicit condition recovered. Representation well-suited to the physics. Mapping covers every qualitative phrase, not just the numeric givens. No errors.
  4 (solid, usable):       Core physics right, but one minor gap -- e.g. a missing implicit constraint, an under-specified representation, a qualitative phrase left un-mapped, or a minor labelling slip. Not a conceptual error.
  3 (partial, default):    Core physical picture is right but has a clear hole -- a missing force, a mis-scoped object list, wrong representation choice for the regime, or an incomplete mapping that would let later [solving] go wrong. START HERE and adjust up or down.
  2 (weak):                A conceptual error (wrong physics law invoked, wrong kinematics assumption, double-counted/missing interaction), or two or more sub-sections are severely incomplete.
  1 (unusable):            Entirely wrong physics, the section is absent, or the content is not a modeling analysis at all.

Think step-by-step internally, then return ONLY valid JSON in exactly this format:
{"reasoning": "<1-3 sentences: what was correct, what was missed, which rubric tier applies>", "score": <integer 1-5>}
Do NOT include markdown, code fences, or any other text outside the JSON object.
\end{tcblisting}

\section{Checker Implementation}
\label{app:checker}

\paragraph{Theorem-library construction.}
The library is built from the PhysReason dataset. We extract physical
laws from its reference solutions, normalize equivalent forms, and remove
duplicates. This process yields 147 reusable templates. The templates contain
general laws and applicability conditions, not problem-specific answers. The
synthesis pipeline instantiates the templates for each problem and validates
the resulting assertions in a sandbox.

For each training problem, we  bind a Python auditor
class, \texttt{PhysicsChecker},  to theorems retrieved from a shared library of 147 physics theorems.
The ground-truth reasoning, target answers, and structured
\texttt{[modeling]} / \texttt{[solving]} fields are used to construct
the problem-specific variable contract and to identify the physical
relations that the checker should audit. The shared theorem library
provides reusable physical-law templates, while the problem-level
reference information instantiates those templates for each problem.
The checker is deliberately unit-test-like: it extracts a fixed variable contract
with missing values represented as \texttt{None}, then runs small independent
tests for interface validity, theorem invariants, trace consistency, and
response under perturbations.
Final-answer matching is treated only as a weak sanity check; the primary reward
signal comes from local physics relations that can be re-derived from the
problem givens.

\paragraph{Synthesis pipeline and cost.}
Checker synthesis is fully automated and cheap. The pipeline has four
stages. It extracts the entities and attributes of a problem, binds them to
theorems from the library, emits the assertions, and validates the resulting
class in the sandbox. No stage requires manual authoring. For the
full PhysReason training set of 1,080 problems, synthesis costs \$5.35 in API
calls in total, about half a cent per problem. Checker creation is therefore
not a scaling bottleneck.

\paragraph{Checker-generation system prompt.}
\begin{tcblisting}{
  longpromptbox,
  title={System prompt for checker synthesis},
}
You are a Senior Physics Auditor. Your mission is to generate a robust Python class `PhysicsChecker`.

### ROBUSTNESS RULES (VITAL):
1. **Safe Attribute Access**: In `__init__`, use `None` as default for ALL missing values:
   ```python
   self.U = getattr(solution_instance, 'U', None)
   # or
   self.U = solution_instance.results.get('U', None)
   ```
   NEVER use numeric defaults like 0.0, 9.81, 2.0 -- let tests detect missing data explicitly.

2. **Fuzzy Numerical Matching (for invariant/modeling tests ONLY)**:
   In `test_invariant_*` and `test_modeling_*` / `test_trace_*` methods, use:
   `math.isclose(a, b, rel_tol=1e-2, abs_tol=1e-2)`
   Physics data via SFT often has rounding noise; be lenient with 1-2% difference.
   NOTE: `math.isclose` is FORBIDDEN in `test_answer_sanity_*` methods.

3. **Force Direction Logic**: When auditing Newton's Laws, carefully analyze the explanation steps and pseudocode.
   For example, if a motor and counterweight both pull a trolley uphill, their tensions are ADDITIVE (T1 + T2).

4. **Symbolic & Numerical Fallback**: Use SymPy for symbolic laws, but always provide a numerical check as a fallback.

5. **Evidence Hierarchy Matters**: Final correctness is determined primarily by the target answers, physics_mapping, key_constraints, theorem knowledge, and explanation_steps. Solving-stage pseudocode is only a reasoning-trace reference and output-contract reference. If pseudocode conflicts with the answer/physics constraints, prefer the latter.

6. **Use Pseudocode Conservatively**: Use solving-stage pseudocode mainly to understand intermediate variable names, [emit] contracts, and trace-consistency checks. Do NOT treat it as the final authority on physics correctness.

7. **Do Not Patch Missing Physics With Defaults**: In `__init__`, do NOT assign numeric fallback defaults like `0.0`, `9.81`, `2.0`, etc. for required givens or result quantities. Use `None` for missing values and let tests fail explicitly with informative messages.

8. **Answer Checks Are Weak Sanity Only**: It is acceptable to include at most one weak `test_answer_sanity_*` method, but it must be secondary. The checker should focus much more on theorem-based sanity checks and modeling quality than on final-answer matching.

9. **Exact Values Are Allowed Only When Derived In-Checker**: It is acceptable to check an exact intermediate/final value if that value is re-derived inside the checker from givens, theorem constraints, or physics relationships. Avoid tests that simply compare against memorized constants copied from the answer key or explanation text.

10. **Do Not Copy Whole Explanation Tables Into One Test**: `explanation_steps` is useful for identifying which relations and substeps matter. It is NOT a license to hardcode a long list of solved constants like `T1=...`, `T2=...`, `sin_theta=...`, `k=...`, `a=...`, `L=...` inside one "consistency" test. One modeling/trace test should check one local relation, not replay the entire standard solution numerically.

11. **Graceful Handling of Missing Data**: Every test method should handle `None` values gracefully:
    ```python
    def test_invariant_example(self):
        if self.some_value is None:
            return (False, "SKIP: some_value not available")
        # ... actual check ...
    ```
    The checker must NEVER crash -- even when given an empty solution_instance with no givens/results.

### CHECKER ARCHITECTURE:
- `__init__(self, solution_instance)`: Extract inputs and results safely using `None` defaults. Define the full variable contract needed by downstream tests.
- The checker should be **unit-test-like**, not one giant monolithic function.
- Define multiple small test methods with these prefixes:
  - `test_interface_*`     -- check that required keys exist and have valid types
  - `test_invariant_*`     -- theorem-based physical sanity (use math.isclose here)
  - `test_modeling_*` or `test_trace_*` -- modeling quality checks
  - `test_response_under_perturbations_*`  -- directional / trend checks via explicit input perturbation
- Optionally define at most one `test_answer_sanity_*` method.
- Each `test_*` method should check exactly one relation or sanity condition and return `(passed_bool, message_str)`.
- `run_all_tests()` must discover all bound methods whose names start with `test_`, execute them one by one IN A TRY/EXCEPT, collect messages, and return `(success_bool, logs_list)`.
  Pattern:
  ```python
  def run_all_tests(self):
      logs = []
      all_pass = True
      for name in sorted(dir(self)):
          if not name.startswith("test_"):
              continue
          method = getattr(self, name)
          if not callable(method):
              continue
          try:
              passed, msg = method()
          except Exception as e:
              passed, msg = False, f"{name} CRASHED: {e}"
          logs.append(f"{'PASS' if passed else 'FAIL'}: {name}: {msg}")
          if not passed:
              all_pass = False
      return (all_pass, logs)
  ```
- The primary emphasis should be:
  - theorem-based physical sanity checks
  - modeling-quality checks based on explanation_steps
  - only then a weak final-answer sanity check
- Use `explanation_steps` to decide what to audit, not to paste solved constants.
- A precise value check is good when it is derived from current givens/results inside the checker.
- A bulk "step consistency" test that hardcodes many solved numbers is low quality and should be avoided.
- Every `test_response_under_perturbations_*` method must:
  - choose one physical input parameter
  - perturb it slightly (for example +1%, +5%, +10%)
  - recompute or estimate the relevant output under that perturbation
  - compare the perturbed output against the baseline output
- A sign check, positivity check, finite-value check, or range check does NOT count as response under perturbations check.
- Prefer many specific tests like `test_interface_required_keys`, `test_invariant_displacement_relation`, `test_modeling_collision_equation`, `test_response_under perturbations_spacing_increases_with_wavelength`.
- Do not collapse all interface checks into one big `test_interface()` method or all invariants into one giant `test_invariants()` method.
- Do not let answer matching dominate the checker. A checker that mostly compares outputs against final answer numbers is low quality.
- If you define `test_answer_sanity_*`, it may only check:
  - sign
  - finite / non-NaN
  - broad magnitude range
  - simple ordering constraints
- `test_answer_sanity_*` must NOT:
  - compute `expected_*`, `target_*`, or `derived_*` values
  - use `math.isclose`
  - solve equations or simplify symbolic expressions
  - re-derive the final answer from formulas

Output ONLY raw Python code with imports (sympy, math, traceback). No markdown fences.
\end{tcblisting}
\paragraph{Example checker fragment.}

The following simplified fragment illustrates numerical checks
for the arc-field problem. 
The attributes \texttt{m}, \texttt{g}, and \texttt{R} are fixed problem
givens; the remaining attributes contain candidate-emitted quantities.
The illustrative interface \texttt{run\_candidate(givens)} executes the
same translated candidate program in a fresh sandbox with the supplied
inputs and returns its emitted quantities as a dictionary.
For each perturbation, the checker preserves the problem's validity
conditions, re-executes the same translated candidate program with the
perturbed inputs, and compares the resulting outputs with the expected
physical response. Each test method returns a Boolean result
and a diagnostic message.

\begin{tcblisting}{
  longpromptbox,
  title={Illustrative checker methods for the arc-field case},
  listing options={
    style=appendixlistingtiny,
    language=Python,
  },
}
import math

def _finite(self, *values):
    return all(
        v is not None and math.isfinite(float(v))
        for v in values
    )

def test_invariants(self):
    """Check candidate outputs against problem-derived relations."""
    try:
        values = (
            self.m, self.g, self.R, self.E_p,
            self.v_O, self.a_x, self.a_y,
        )
        if not self._finite(*values):
            return False, "Missing or non-finite quantity"

        if min(self.m, self.g, self.R) <= 0 or self.v_O <= 0:
            return False, "Invalid givens or velocity branch"

        # Problem givens: v_B^2 = gR, qE = sqrt(2) mg,
        # and the electric force points at 45 degrees.
        energy_scale = self.m * self.g * self.R

        # On the arc: W_g = -mgR, W_E = 2mgR.
        # Compare this work with the candidate's kinetic-energy change.
        delta_ke = 0.5 * self.m * (
            self.v_O**2 - self.g * self.R
        )

        # Normalize residuals to obtain dimensionless tolerances.
        # Assertions are grouped here for compact illustration.
        checks = {
            "spring energy":
                abs(self.E_p / energy_scale - 0.5) <= 1e-2,
            "arc work-energy":
                abs(delta_ke / energy_scale - 1.0) <= 1e-2,
            "horizontal acceleration":
                abs(self.a_x / self.g - 1.0) <= 1e-2,
            "vertical acceleration":
                abs(self.a_y / self.g) <= 1e-2,
        }

        failed = [
            name for name, passed in checks.items() if not passed
        ]
        message = "; ".join(failed) if failed else "Invariants satisfied"
        return not failed, message

    except Exception as exc:
        return False, f"Invariant check failed: {exc}"


def test_trace_consistency(self):
    """Check consistency among candidate-emitted quantities."""
    try:
        values = (
            self.x, self.y, self.v_O,
            self.a_x, self.a_y, self.R,
        )
        if not self._finite(*values):
            return False, "Missing or non-finite trace quantity"

        if (
            self.R <= 0 or self.v_O <= 0
            or self.x < 0 or self.y < 0
        ):
            return False, "Invalid scale or trajectory branch"

        # At O: x(0) = y(0) = 0, vx(0) = 0, vy(0) = v_O.
        # Use the candidate's x and a_x to infer elapsed time,
        # then check its reported y against its own motion parameters.
        if self.a_x <= 0:
            return False, "Expected positive horizontal acceleration"

        t = math.sqrt(2.0 * self.x / self.a_x)
        y_from_trace = (
            self.v_O * t + 0.5 * self.a_y * t**2
        )

        length_scale = max(
            self.R, abs(self.y), abs(y_from_trace)
        )
        residual = abs(self.y - y_from_trace) / length_scale

        passed = residual <= 1e-2
        message = (
            "Trace consistent" if passed
            else "Inconsistent trajectory"
        )
        return passed, message

    except Exception as exc:
        return False, f"Trace check failed: {exc}"


def test_response_under_perturbations(self):
    """Re-execute the same candidate at two physically valid radii."""
    try:
        if not self._finite(self.m, self.g, self.R):
            return False, "Missing or non-finite givens"

        if min(self.m, self.g, self.R) <= 0:
            return False, "Invalid givens"

        radii = (self.R, 1.10 * self.R)
        velocities = []

        for radius in radii:
            # Preserve the problem's defining constraints when R changes.
            givens = {
                "m": self.m,
                "g": self.g,
                "R": radius,
                "v_B": math.sqrt(self.g * radius),
                "qE": math.sqrt(2.0) * self.m * self.g,
                "theta": math.pi / 4.0,
            }

            # Execute the fixed candidate program, not a reference formula.
            result = self.run_candidate(givens)
            velocity = result.get("v_O")

            if (
                not self._finite(velocity)
                or float(velocity) <= 0
            ):
                return False, "Missing or invalid candidate velocity"

            velocities.append(float(velocity))

        v0, v1 = velocities

        # With the stated constraints preserved, v_O scales as sqrt(R).
        # Compare candidate responses with this physical scaling relation.
        observed_ratio = v1 / v0
        expected_ratio = math.sqrt(radii[1] / radii[0])

        passed = (
            v1 > v0
            and math.isclose(
                observed_ratio,
                expected_ratio,
                rel_tol=1e-2,
                abs_tol=0.0,
            )
        )

        message = (
            "Response consistent" if passed
            else "Incorrect response to R"
        )
        return passed, message

    except Exception as exc:
        return False, f"Perturbation check failed: {exc}"
\end{tcblisting}

\section{Checker Reliability Diagnostics}
\label{app:checker_diag}

The checker supplies a denser and largely reliable signal. Its residual
errors fall into a few identifiable failure types. To show this, we
reconstructed every GRPO group formed during RLAWM training on PhysReason with
Qwen3-VL-8B-Instruct. We then analyzed the rewards from three angles: reward
sparsity, the causes of $R_S=0$, and disagreement with the final answer. The
three analyses below support one conclusion: the checker’s disagreements with final-answer correctness are
measurable and arise from identifiable coverage and interface issues.

\paragraph{Reward sparsity.}
The checker adds gradient exactly where the outcome reward carries none. The
GRPO gradient comes from within-group relative rewards, so a group whose 8
rollouts score identically contributes nothing. A checker runs 7.34 assertions
per rollout on average, so $R_S$ takes graded values rather than a binary one.
Table~\ref{tab:sparsity} shows
that 29.9\% of groups are in this state under the outcome-only reward. Adding
the checker reduces this share to 9.3\%, recovering 69.0\% of the zero-variance
groups. Among all-wrong rollouts, 64.1\% still receive a nonzero $R_S$. Hard
problems whose rollouts all fail therefore re-enter training instead of being
silently dropped.

\paragraph{Why rollouts receive $R_S=0$.}
Zero solving scores stem mainly from execution mechanics rather than from the
checker misjudging physics. A rollout is scored $R_S=0$ in 19.4\% of cases.
Table~\ref{tab:checker_failures} breaks this down by cause. Sandbox execution
errors dominate, and variable-contract mismatch is the second source. Contract
mismatch means the pseudocode compiles but some freely named variables cannot
be bound to the checker's fields. Translation failure and missing solving code
together account for under one percent. The zero scores are therefore a bounded
cost of executing the trace, not a sign of unreliable physics judgment.

\begin{table}[h]
\centering
\small
\caption{\textbf{Failure types behind $R_S=0$.} Share of all training rollouts on PhysReason with Qwen3-VL-8B-Instruct.}
\label{tab:checker_failures}
\begin{tabular}{@{}l c@{}}
\toprule
Failure type & Share of all rollouts \\
\midrule
Sandbox execution error & 11.8\% \\
Variable-contract mismatch & 7.0\% \\
Pseudocode translation failure & 0.4\% \\
Missing solving code & 0.2\% \\
\midrule
Total (scored $R_S=0$) & 19.4\% \\
\bottomrule
\end{tabular}
\end{table}

\paragraph{Agreement and disagreement with final-answer correctness.}
The checker's disagreements with the final answer are asymmetric: 4.9\% false
negatives against 15.8\% false positives. Among rollouts with a correct answer
and successful execution, 89.7\% receive full credit, 5.4\% partial credit, and
4.9\% are fully rejected. The rejections have three sources: contract gaps,
legitimate derivation structures not covered by the synthesized tests, and
overly strict assertions. False positives occur when the error lies outside the
checker's coverage. The most common case is a wrong final algebraic
simplification after all audited relations hold. The low false-negative rate
means valid alternative derivations are rarely punished. The false-positive rate
is why the joint reward in \S\ref{sec:joint_objective} retains the final-answer
term $R_A$.

\paragraph{Failure cases.}
Two cases show where the checker's judgment ends. The first is a false
negative. A rollout solves a three-step gas cycle correctly under the
work-done-on-gas sign convention, yet loses almost half of $R_S$. The second is
a false positive. A rollout solves a series RLC resonance problem with every
audited relation intact, yet writes a wrong bandwidth in its final answer. One
case under-rewards correct physics, the other over-rewards a wrong final line.

\subsubsection*{Failure case 1: false negative (all sub-answers correct, $R_S = 0.56$)}

\begin{tcolorbox}[
  appendixbox,
  title={Problem (three-step gas cycle)},
]
One mole of an ideal gas with adiabatic index $\gamma$ starts at
$(P_0, V_0, T_0)$. Step 1: adiabatic compression to volume $V_0/r$. Step 2:
isobaric expansion back to $V_0$. Step 3: isochoric cooling back to the initial
state.

\medskip
\noindent\textbf{Sub-questions.}
\textbf{(Q1)} the work done by the gas in step 1;
\textbf{(Q2)} the heat absorbed in step 2;
\textbf{(Q3)} the efficiency of the cycle.
\end{tcolorbox}

\begin{tcblisting}{
  longpromptbox,
  title={Rollout, condensed to its decisive lines},
  listing options={
    style=appendixlistingtiny,
    basicstyle=\ttfamily\tiny,
    escapeinside={(*}{*)},
  },
}
[modeling]
[constraints]
(*\textbf{- first law written as dU = Q + W, with W the work done ON the gas (used consistently below)}*)
...
[solving]
[sub1]
- adiabatic:  T_1 = T_0 r^(gamma-1),   P_1 = P_0 r^gamma                                 -> emit T_1, P_1
- Q_1 = 0;   W_1 = dU_1 = P_0 V_0 (r^(gamma-1) - 1)/(gamma-1)    [W_1 > 0: work on gas]  -> emit Q_1, W_1, dU_1
[sub2]
- isobaric:  T_2 = T_1 r;   W_2 = -P_1 (V_0 - V_0/r) = -P_0 V_0 (r^gamma - r^(gamma-1))   -> emit T_2, W_2
- Q_2 = dU_2 - W_2 = (gamma/(gamma-1)) P_0 V_0 (r^gamma - r^(gamma-1))                    -> emit Q_2, dU_2
[sub3]
- isochoric:  W_3 = 0;   Q_3 = dU_3 = -P_0 V_0 (r^gamma - 1)/(gamma-1)                    -> emit W_3, Q_3, dU_3
- W_by = -(W_1 + W_2 + W_3);   eta = W_by / Q_2                                         -> emit W_by, eta

[ans]
work by the gas in step 1 = -W_1 = P_0 V_0 (1 - r^(gamma-1))/(gamma-1);   Q_2 as above;
eta = 1 - (r^gamma - 1) / (gamma (r^gamma - r^(gamma-1)))
(*\textbf{All three sub-answers are correct.}*)
(*\textbf{The translator binds W\_1, W\_2 to the checker's work-done-BY-gas fields; each test that reads a work term sees the opposite sign.}*)
\end{tcblisting}

\paragraph{Why the checker rejects it.}
The rollout is under-rewarded for a sign convention, not for its physics.
The checker writes the first law as $\Delta U = Q - W$, with $W$ the work done
by the gas. The rollout writes it as $\Delta U = Q + W$, with $W$ the work done
on the gas, and uses this form consistently. The translator binds the rollout's
$W_i$ to the checker's $W_i$ fields, so every test that reads a work term sees
the opposite sign. Four of nine tests do: the two first-law tests, the
isobaric-work trace test, and the net-work perturbation test. These four fail
and the other five pass, giving $R_S = 0.56$. The penalty stays bounded because
the tests are independent and the convention touches only some of them. When
most tests of a problem read the affected quantity, the same mechanism rejects
the rollout outright. This case combines two false-negative sources from the
main analysis. The assertion fixes a convention that the law leaves open, and
the bound field carries a meaning the rollout did not intend. The law under
test is the same on both sides. Only its written form differs.

\subsubsection*{Failure case 2: false positive ($R_S = 1.0$, one sub-answer wrong)}

\begin{tcolorbox}[
  appendixbox,
  title={Problem (series RLC resonance)},
]
A resistor $R$, an inductor $L$, and a capacitor $C$ are connected in series
and driven by $v(t) = V_0\cos\omega t$.

\medskip
\noindent\textbf{Sub-questions.}
\textbf{(Q1)} the steady-state current amplitude $I_0(\omega)$ and phase
$\varphi$;
\textbf{(Q2)} the average power $\langle P\rangle(\omega)$ and its maximum;
\textbf{(Q3)} the half-power angular frequencies $\omega_\pm$ and the bandwidth
$\Delta\omega = \omega_+ - \omega_-$.
\end{tcolorbox}

\begin{tcblisting}{
  longpromptbox,
  title={Rollout, condensed to its decisive lines},
  listing options={
    style=appendixlistingtiny,
    basicstyle=\ttfamily\tiny,
    escapeinside={(*}{*)},
  },
}
[solving]
[sub1]
- X = omega L - 1/(omega C),  Z = sqrt(R^2 + X^2),  I_0 = V_0/Z,  phi = arctan(X/R)   -> emit X, Z, I_0, phi
[sub2]
- P_avg = V_0^2 R / (2 Z^2);  at omega_0 = 1/sqrt(L C):  P_max = V_0^2 / (2 R)        -> emit P_avg, omega_0, P_max
[sub3]
- half power:  Z^2 = 2 R^2   <=>   X = +R  or  X = -R
- w_plus  = +R/(2L) + sqrt(R^2/(4L^2) + 1/(L C))                                      -> emit w_plus
- w_minus = -R/(2L) + sqrt(R^2/(4L^2) + 1/(L C))                                      -> emit w_minus
- dw = w_plus - w_minus                                                              -> emit dw
[subans 3]
omega_pm = +/- R/(2L) + sqrt(R^2/(4L^2) + 1/(L C));   (*\textcolor{red!70!black}{\textbf{Delta omega = w\_plus - w\_minus = R/(2L)}}*)
(*\textbf{The two R/(2L) terms add to R/L. The emitted dw is correct; the written closed form is not.}*)

[ans]
I_0 = V_0/Z, phi = arctan(X/R);   P_max = V_0^2/(2R);   omega_pm = ...,  (*\textcolor{red!70!black}{Delta omega = R/(2L)}*)
(*\textbf{Sub-answer 3 is wrong.}*)
\end{tcblisting}

\paragraph{Why the checker passes it.}
The rollout passes because its error lies outside the emitted variable
contract. Every emitted quantity is correct, including the two half-power
roots and their difference \texttt{dw}. The slip happens only when the rollout
writes the closed form of $\Delta\omega$ into its sub-answer. The two
$\pm R/(2L)$ terms should add to $R/L$, but the rollout writes $R/(2L)$. The
checker audits \texttt{dw}, not the sub-answer text. Its answer-sanity test
checks only sign, finiteness, and magnitude range. It may not simplify or
compare expressions, so $R/(2L)$ passes. The perturbation tests check trends,
and a constant factor leaves every trend intact. The physics is sound, so a
full $R_S$ is defensible. The penalty comes from $R_A$ instead. This is the most
common false-positive pattern in the main analysis.

\paragraph{What the two cases imply.}
The two cases mark the two edges of checker coverage. On one edge, a law
test carries a written form, so a rollout that obeys the law under another
convention is under-rewarded. This motivates convention-invariant tests and the
adaptive probe of \S\ref{sec:conclusion}. On the other edge, what the rollout
writes after its last emit is invisible to the checker. This is why the joint
reward keeps $R_A$.
\section{Baseline Implementation Details}
\label{app:baselines}

The LLM-as-a-judge baselines use the same GRPO optimization
hyperparameters (Appendix~\ref{app:hyperparams}) as RLAWM. They evaluate the complete student solution
with a holistic rubric. The grader’s five-point judgment becomes a single
scalar reward for the whole response.

We use two prompt variants.
\textbf{Direct-Likert} evaluates the student solution without access to the
reference answer or reference reasoning.
\textbf{Reference-Likert} additionally receives the reference final answer(s)
and, when available, reference reasoning steps. Both variants use the same five-point scoring rubric
and request a JSON object containing a brief justification and an
integer score in $\{1,\ldots,5\}$. The reference-based prompt explicitly
allows valid derivations that differ from the reference.

The prompt templates are shown below. Braced placeholders are replaced
with the corresponding example content. For Reference-Likert, the reference
reasoning block is omitted when reference reasoning is unavailable.

\paragraph{Direct-Likert prompt.}
\begin{tcblisting}{
  promptbox,
  title={Direct-Likert judge prompt (no reference)},
  listing options={
    style=appendixlisting,
    language={},
  },
}
You are a process-reward physics grader. You do NOT have access to the ground-truth answer. First, provide a brief reasoning comment about the student's solution, then assign a score from 1 to 5.

Scoring guidelines (1-5):
- 1: Nonsense, violates physical laws, internally inconsistent, unreadable
- 2: Weak reasoning, some correct principles but major errors or gaps
- 3: Moderate reasoning, mostly correct laws applied, minor mistakes, partially clear
- 4: Strong reasoning, correct laws, mostly consistent derivations, minor algebra/unit issues
- 5: Excellent reasoning, fully consistent with physics principles, correct units, plausible intermediate results, clear and logical steps

Focus on:
- Correct application of physical laws
- Logical consistency of derivations and steps
- Correct usage of formulas, symbols, and units
- Reasonable assumptions and approximations
- Physical plausibility of intermediate and final results

Student solution:
{solution}

Return JSON exactly in the following format:
{"reasoning": "...your brief reasoning comment...", "score": x}
\end{tcblisting}

\paragraph{Reference-Likert prompt.}
\begin{tcblisting}{
  promptbox,
  title={Reference-Likert judge prompt (with reference answer and reasoning)},
  listing options={
    style=appendixlisting,
    language={},
  },
}
You are a physics grader. Use the reference solution to evaluate the student's work. First, provide a brief reasoning comment about the student's solution, then assign a score from 1 to 5.

Scoring guidelines (1-5):
- 1: Nonsense, violates physical laws, internally inconsistent, unreadable
- 2: Weak reasoning, some correct principles but major errors or gaps
- 3: Moderate reasoning, mostly correct laws applied, minor mistakes, partially clear
- 4: Strong reasoning, correct laws, mostly consistent derivations, minor algebra/unit issues
- 5: Excellent reasoning, fully consistent with physics principles, correct units, plausible intermediate results, clear and logical steps

Reward the student for:
- Sound reasoning even if the path differs from the reference
- Correct derivations, formula usage, and unit consistency
- Clear explanation of assumptions, approximations, and physical laws applied
- Plausible intermediate results
- Proper handling of edge cases and limiting behaviors

Reference answer(s):
{reference_answers}

Reference reasoning steps:
{reference_reasoning}

Student solution:
{solution}

Return JSON exactly in the following format:
{"reasoning": "...your brief reasoning comment...", "score": x}
\end{tcblisting}

% \newpage
% \input{checklist.tex}

\end{document}